\documentclass[sigconf]{acmart}
\AtBeginDocument{%
  \providecommand\BibTeX{{%
    \normalfont B\kern-0.5em{\scshape i\kern-0.25em b}\kern-0.8em\TeX}}}

\setcopyright{iw3c2w3}
\copyrightyear{2021}
\acmYear{2021}
\acmDOI{10.1145/3442381.3449795}

\acmConference[WWW '21]{Proceedings of the Web Conference 2021}{April 19--23, 2021}{Ljubljana, Slovenia}
\acmBooktitle{Proceedings of the Web Conference 2021 (WWW '21), April 19--23, 2021,
Ljubljana, Slovenia}
\acmPrice{}
\acmISBN{978-1-4503-8312-7/21/04}

\usepackage{bm}
\usepackage{amstext}
\usepackage{amsmath}
\usepackage{amsfonts}
\usepackage{amsthm}
\usepackage{amssymb}

\usepackage{booktabs} 
\usepackage{graphicx}
\usepackage{subcaption}
\usepackage{threeparttable}
\usepackage{algorithm}
\usepackage{algpseudocode}
\usepackage{enumitem}
\usepackage{CJKutf8}
\usepackage{bigstrut}
\usepackage{multirow}

\newcommand{\q}{\bm{q}}
\newcommand{\m}{\bm{m}}
\newcommand{\W}{\bm{W}}

\newcommand{\rto}{\leftarrow}
\newcommand{\bn}{\mathcal{N}}
\newcommand{\be}{\mathcal{E}}
\newcommand{\bs}{\mathcal{S}}
\newcommand{\bd}{\mathcal{D}}
\newcommand{\bu}{\mathcal{U}}
\newcommand{\bv}{\mathcal{V}}
\newcommand{\bq}{\mathcal{Q}}
\newcommand{\usu}{\text{USU}}
\newcommand{\dsd}{\text{DSD}}

\newcommand{\npmi}{\text{npmi}}

\newsavebox\CBox
\def\textBF#1{\sbox\CBox{#1}\resizebox{\wd\CBox}{\ht\CBox}{\textbf{#1}}}

\begin{document}

\title{Online Disease Diagnosis with Inductive Heterogeneous Graph Convolutional Networks}

\author{Zifeng Wang}
\email{wangzf18@mails.tsinghua.edu.cn}
\affiliation{%
  \institution{TBSI, Tsinghua University}
}

\author{Rui Wen}
\email{ruiwen@tencent.com}
\affiliation{%
  \institution{Tencent Jarvis Lab}
}

\author{Xi Chen}
\authornote{Corresponding author.}
\email{jasonxchen@tencent.com}
\affiliation{%
  \institution{Tencent Jarvis Lab}
}

\author{Shilei Cao}
\email{eliasslcao@tencent.com}
\affiliation{%
  \institution{Tencent Jarvis Lab}
}

\author{Shao-Lun Huang}
\email{shaolun.huang@sz.tsinghua.edu.cn}
\affiliation{%
  \institution{TBSI, Tsinghua University}
}

\author{Buyue Qian}
\email{qianbuyue@xjtu.edu.cn}
\affiliation{%
  \institution{Xi'an Jiaotong University}
}

\author{Yefeng Zheng}
\email{yefengzheng@tencent.com}
\affiliation{%
  \institution{Tencent Jarvis Lab}
}


\begin{abstract}
We propose a Healthcare Graph Convolutional Network (HealGCN) to offer disease self-diagnosis service for online users based on Electronic Healthcare Records (EHRs). Two main challenges are focused in this paper for online disease diagnosis: (1) serving cold-start users via graph convolutional networks and (2) handling scarce clinical description via a symptom retrieval system. To this end, we first organize the EHR data into a heterogeneous graph that is capable of modeling complex interactions among users, symptoms and diseases, and tailor the graph representation learning towards disease diagnosis with an inductive learning paradigm. Then, we build a disease self-diagnosis system with a corresponding EHR Graph-based Symptom Retrieval System (GraphRet) that can search and provide a list of relevant alternative symptoms by tracing the predefined meta-paths. GraphRet helps enrich the seed symptom set through the EHR graph when confronting users with scarce descriptions, hence yield better diagnosis accuracy. At last, we validate the superiority of our model on a large-scale EHR dataset.
\end{abstract}



\maketitle

\section{Introduction}
Electronic Health Records (EHRs) are documented information on many clinical events that occur during a patient's stay and visit in the hospital. Recently, the advancement of machine learning sheds light on building an alternative substitute ``robot doctor'', which benefits from massive EHR data and empirically learns a disease diagnosis model based on the growing collections of clinical observations. Automatic disease diagnosis can benefit a lot in the current medicine system. For example, primary care providers, who are responsible for coordinating patient care among specialists and other care levels, often confront a mix of diseases which should belong to various departments. In this circumstance, they could refer to the automatic online disease diagnosis results as useful advice and decide which level of healthcare service patients would need. Moreover, users can also get access to this online diagnosis system for getting advice based on their chief complaints, then decide which hospital and department they would like to go for a visit. In a nutshell, an advanced automatic online diagnosis system can boost the efficiency of the existing medicine system.

Some documented EHRs were recently made publicly available, e.g., the MIMIC-III \cite{johnson2016mimic} and CPRD \cite{herrett2015data}, which encourage a surge of research in developing automatic clinical expert systems. Existing works in utilizing EHRs for disease diagnosis include three genres: feature engineering associated with a classifier for the aim of predicting disease outcomes \cite{che2017deep,farhan2016predictive,purushotham2017benchmark,wang2021lifelong}; unsupervised representation learning based on raw notes \cite{choi2016multi,choi2018mime, li2019behrt} by modeling the semantic relations between diseases and symptoms for downstream tasks; and graph-based methods \cite{hosseini2018heteromed, hosseini2019hierarchical} by modeling the EHR data with graphs, then attempting to tailor embedding learning specifically for disease diagnosis. In this work, we follow the idea of modeling EHR data into Heterogeneous Information Network (HIN) \cite{han2010mining}. We perform graph representation learning with a link prediction paradigm by Graph Convolutional Networks (GCNs) \cite{kipf2016semi}. HIN is inherently suitable for modeling the complex interactions among users, symptoms and diseases in the EHR data. It allows to analyze the diagnosis results conveniently by tracing the links between nodes of symptoms and diseases through an interpretable way.
\begin{figure}[t]
\centering
\includegraphics[width=0.40\textwidth]{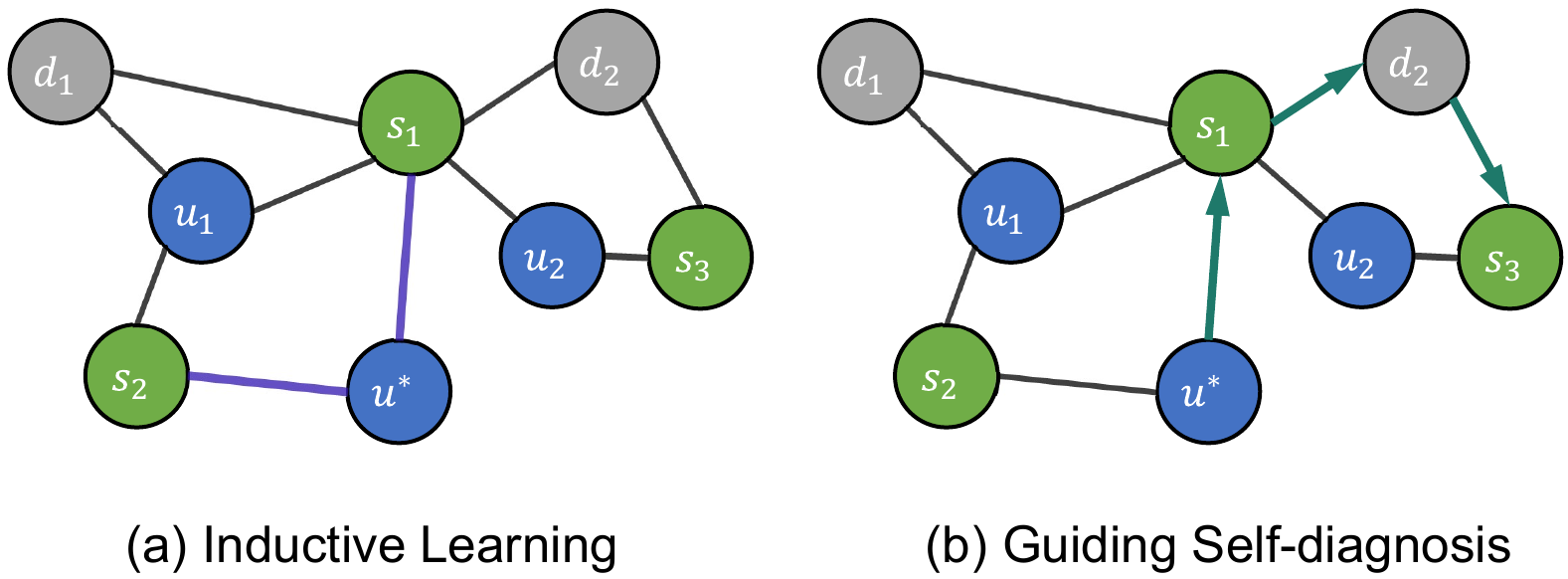}
\caption{Demonstration of (a) inductive learning that a patient $u^*$ needs to be incorporated into the graph and (b) guiding self-diagnosis by exploring possible symptoms through meta-path. Here, $u$, $s$ and $d$ represent the user, symptom, and disease node, respectively, in a Heterogeneous Information Network (HIN). \label{fig:1}}
\end{figure}

It should be noted that our work emphasizes on \textbf{disease diagnosis} rather than \textbf{individual disease risk prediction}, such that it differs from most previous deep learning based methods, e.g., DoctorAI \cite{choi2016doctor}, RETAIN \cite{choi2016retain}, Dipole \cite{ma2017dipole}, etc.  They leverage recurrent neural network (RNN) to model \emph{sequential} EHR data by examining each user's historical visits for predicting his/her future disease risk. Nonetheless, all of them are \emph{transductive} and unable to deal with \emph{cold-start} users, i.e., users do not have historical records of hospital visits. 

Unfortunately, in online self-diagnosis, we often have \textbf{little prior knowledge} about individuals compared with those in the public MIMIC-III or CPRD datasets. Users access to a web-based diagnosis system where neither personal information is collected nor their previous logs are found. The system should make a decision purely based on questioning and answering with newcomers. It could be expected information provided by a single user is far from enough to make an accurate and confident diagnosis. Such that, exploiting the HIN to get information from the neighbors should benefit a lot. However, recent graph based disease risk prediction methods \cite{hosseini2018heteromed, hosseini2019hierarchical, hettige2019medgraph} all fall short in transductive learning paradigm, which rely on user's hospital visits involving the detailed clinical notes written by physicians, physical examination, etc., thus not applicable to online self-diagnosis.

To deal with the above-mentioned challenge, we propose to employ \emph{inductive learning} \cite{hamilton2017inductive}. A new patient's embedding is generated by inductive graph convolutional operation tracing the predefined meta-path. This process encodes the node information by exploiting the high-order connectivity. As shown in Fig. \ref{fig:1}(a), a new patient node $u^*$ is incorporated in the existing EHR graph by its connection with two symptom nodes $s_1$ and $s_2$. We can generate embedding of $u^*$ by an aggregation operation from its two neighbors, e.g., taking average of the embeddings of $s_1$ and $s_2$. Moreover, we can involve neighbors of $s_1$ and $s_2$ into aggregation, e.g., users $u_1$ and $u_2$ who are so-called two-hop neighbors of $u^*$ with respect to the meta-path user-symptom-user.

The scarcity of clinical descriptions is another obstacle in developing an accurate self-diagnosis system for practical use, but has long been ignored in the literature. Notes by physicians, which synthetically reflect patients' physical conditions, are core ingredients of former models \cite{hosseini2018heteromed, hosseini2019hierarchical}. However, without professional knowledge in medicine, an ordinary patient cannot provide accurate descriptions about symptoms but could merely present a \emph{colloquial description} (such as high temperature and feel sick) to the target symptom, which might be distinctively different in the clinical definition. In this circumstance, we leverage an named entity recognition (NER) system to extract key symptoms from user inputs, then \textbf{guide} users to offer more symptoms by exploiting the learned HIN, as shown in Fig. \ref{fig:1}(b). According to the meta-path symptom-disease-symptom, we can find out that symptom $s_3$ is a two-hop neighbor of $u^*$, starting from $s_1$ which is directly connected to $u^*$.  As far as we know, few works have been devoted to it in the literature.

In summary, our main contributions are highlighted as:
\begin{itemize}[leftmargin=*, itemsep=0pt, labelsep=5pt]
\item We propose HealGCN, an inductive heterogeneous GCN-based disease diagnosis model, towards serving cold-start users by mining complex interactions in the EHR data.

\item We build a synthetic disease self-diagnosis system based on HealGCN, medical NER system, and symptom retrieval system for serving online users.

\item We verify the effectiveness of the proposed HealGCN and the corresponding symptom retrieval system on real-world EHR data and in online A/B test.
\end{itemize}

\begin{figure}[t]
\centering
\includegraphics[width=0.5\textwidth]{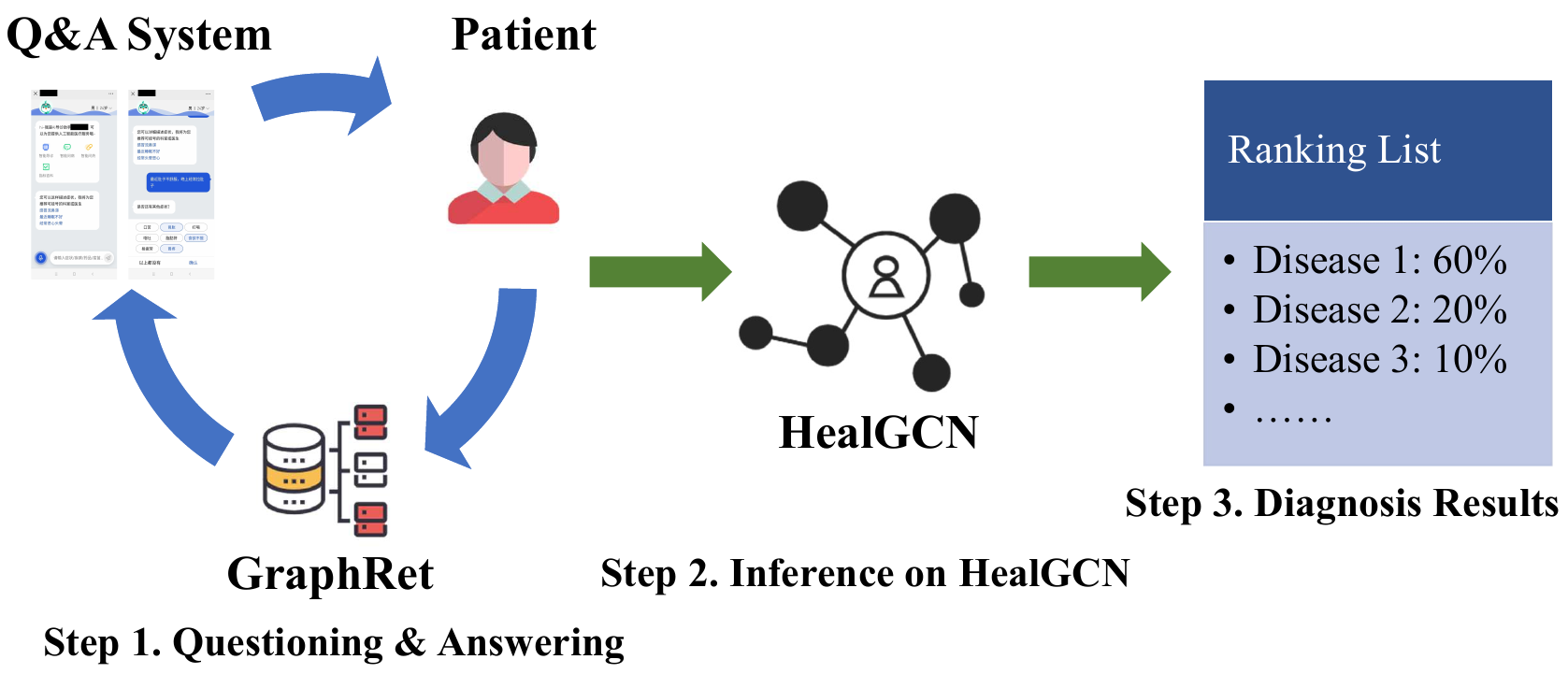}
\caption{The flowchart of the automatic diagnosis system. \label{fig:4}}
\end{figure}

\section{Related Work}
A series of works have been devoted to disease diagnosis by machine learning. Traditional wisdom focused on feature engineering with domain knowledge, along with statistical classification models for disease prediction \cite{lin2009intelligent,soni2011predictive,weng2016disease,lixian2021}. These methods often require intensive labor in data preprocessing and professional knowledge to design discriminative handcrafted features. Inspired by the emergence of powerful techniques in natural language processing (NLP), e.g., Word2Vec \cite{mikolov2013efficient}, many followups were proposed to mine clinical notes through an unsupervised word representation learning scheme, in order to support the downstream disease prediction task engaged with a multi-class classification model \cite{choi2016learning,choi2016multi, farhan2016predictive,mullenbach2018explainable}. Recent advancement in pretrained language models, e.g., BERT \cite{devlin2018bert}, further fueled research in formulating EHR data parsing as text processing \cite{li2019behrt}. However, these self-supervised learning methods only learn general embeddings, which are  unnecessarily optimal for disease diagnosis. Besides, they ignore the rich structural information of EHR data, e.g., the interactions among users, symptoms and diseases, which should be useful to build accurate and interpretable diagnosis systems.

\begin{figure*}[t]
\centering
\includegraphics[width=0.9\textwidth]{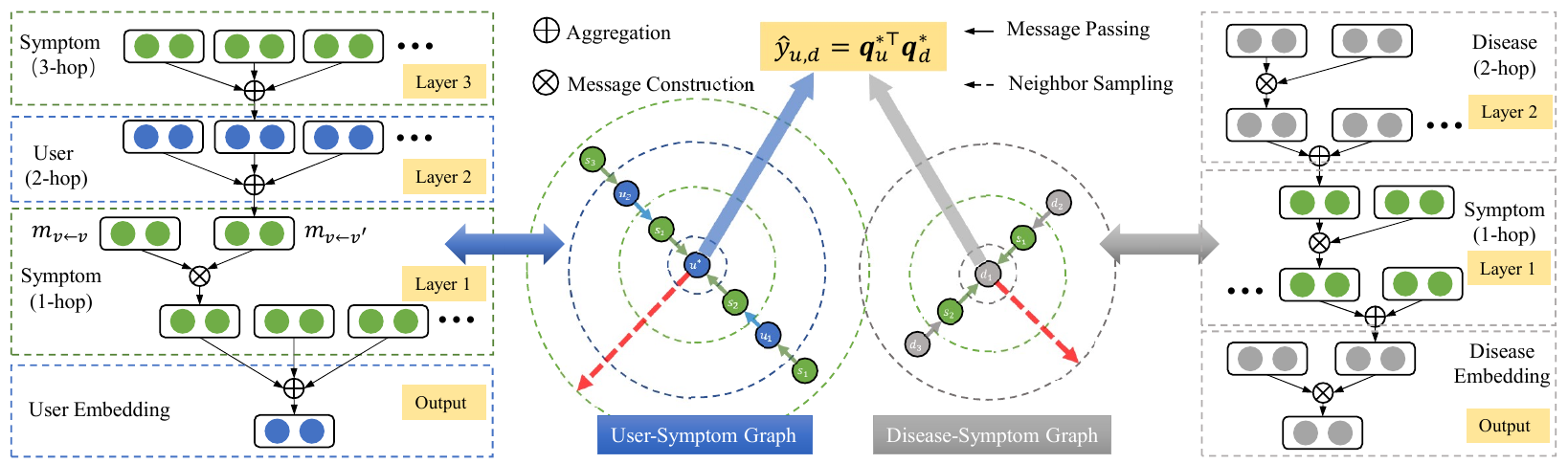}
\caption{Overall flowchart of the proposed HealGCN framework. \label{fig:2}}
\end{figure*}

In order to utilize rich semantics contained in EHR data, HeteoMed \cite{hosseini2018heteromed} proposed HIN-based graph neural networks targeting for disease diagnosis, followed by works in adopting Graph Convolutional Transformer (GCT)  \cite{choi2019graph}  and attention GCN \cite{hosseini2019hierarchical}. However, these methods are not capable of handling cold-start users because they model patient embeddings based on abundant historical clinical events, thus being not able to handle patients without historical visits. In this work, we leverage an inductive heterogeneous GCN to resolve this challenge.

There were a lot of research works on the vectorization of clinical concepts, including patients \cite{zhu2016measuring, zhang2018patient2vec,dligach2018learning}, doctors \cite{biswal2019doctor2vec} and medical notes \cite{galko2018biomedical, wei2018embedding}, for clinical information retrieval. However, many of them rely on deep architectures for extracting embeddings for pair-wise comparison, which are not easy to scale up to a large number of alternative symptoms, or are not targeting specifically for symptom-level retrieval. On contrast, we exploit meta-path in the EHR graph with GCNs for generating symptom embeddings, which are then exploited for symptom retrieval in the self-diagnosis question and answer (Q\&A) system. Our retrieval system takes disease as intermediate node to bridge symptoms, which constraints the search to more relevant symptoms and increases the retrieval diversity.

\section{Method} \label{sec:3}
The overall flowchart of the proposed automatic diagnosis system is shown by Fig. \ref{fig:2}. It encompasses three main steps: question and answer, inference for diagnosis, and diagnosis results display. In this section, we present technical details of the inference component. The first Q\&A component will be introduced in $\S$ \ref{sec:selfdiag}. We first introduce how to build an HIN based on the EHR data and adapt the graphical reasoning for disease diagnosis. After that, we elaborate on the details of neighbor sampling through the meta-path, and the embedding propagation and aggregation process, on the basis of message functions. At last, we present the idea of employing a contrastive loss function for optimization and utilizing hard negative example mining for long-tailed distribution.

\subsection{Building the EHR HIN Graph}
A \emph{homogeneous} graph is denoted by $\mathcal{G}=(\mathcal{V},\mathcal{E})$ that consists of two elements: node $v \in \mathcal{V}$ and edge $e=(v,v^\prime) \in \mathcal{E}$. By contrast, our graph built on EHR data is \emph{heterogeneous}, i.e., there are three classes of nodes, representing symptom $s$, user $u$, and disease $d$, respectively, thus three types of edges: $(u,s)$, $(u,d)$ and $(d,s)$. These three edge types represent common interactions in the EHR data. In particular, edges $(u,s)$ and $(u,d)$ exist when user $u$ reports symptom $s$ and is affected by disease $d$, and $(d,s)$ reflects that disease $d$ appears together with symptom $s$, according to observations in the EHR data. The node set of our EHR graph is a combination of users, symptoms and diseases, as $\mathcal{V} = \{\mathcal{U},\mathcal{S},\mathcal{D}\}$. Besides, the neighborhood of node $v$ is $\bn(v)=\{v^\prime \in \bv \mid (v,v^\prime) \in \be \}$, from which we will propose the meta-path guided reasoning in the sequel.

\subsection{Problem Setup}
Different from many previous GCN works that try to model disease diagnosis as a node classification task, we formulate the disease diagnosis process as \emph{link prediction} between user nodes $\bu$ and disease nodes $\bd$. Similar to the standard collaborative filtering scheme, finding the most possibly linked disease $d$ with the user $u$ amounts to solving the following problem
\begin{equation}
d = \mathop{\arg\max}_{d^\prime \in \bd}\hat{y}_{u,d^\prime}= \q_u^{\top}\q_{d^\prime},
\label{eq:getscore}
\end{equation}
where $\q$ is the embedding of a node. Here, $\hat{y}_{u,d}$ is the predicted score that measures how possible user $u$ is affected by disease $d$, hence we can deploy top-$K$ disease ``recommendations''  in our self-diagnosis system for users, based on the ranked scores.

However, we do not directly assign an embedding $\q_u$ for each user; instead, we only maintain a trainable embedding matrix of symptoms and diseases:
\begin{equation}
\mathcal{Q}=  [\underbrace{\q_{d·_1},\dots,\q_{d_{|\bd|}}}_{\text{diseases embeddings}}, \underbrace{\q_{s_1},\dots, \q_{s_{|\bs|}}}_{\text{symptoms embeddings}}].
\end{equation}
The main reason is that we have limited knowledge about individuals in online self-diagnosis. For instance, most of users are cold-start, and the number of users can be very large compared with symptoms and diseases. In this scenario, maintaining embeddings of users causes difficulty in optimization on those tremendous number of parameters. Instead, we propose to represent users by the linked symptoms. By exploiting the meta-path methods, we mine the users' position and the structure of their neighborhood to build the embeddings, which reduces the optimization complexity significantly.

\subsection{Meta-Path Guided Neighbor Sampling}
In order to fulfill the potential of the EHR graph for representation learning, unlike \citet{hosseini2018heteromed} that obtains embeddings by aggregation on one-hop neighbors, we exploit high order connections by meta-path. We include two meta-paths in our model: Disease-Symptom-Disease (DSD) and User-Symptom-User (USU). Technically, given a meta-path $\rho$, we define the $i$-hop neighborhood of a node $v$ as $\bn_\rho^{i}(v)$. For instance, in Fig. \ref{fig:1}(a), when exploiting neighbors of $u^*$ in line with the USU meta-path, we can obtain neighbors as $\bn_{\usu}^{1}(u^*)=\{s_1,s_2\}$, $\bn_{\usu}^2(u^*)=\{u_1,u_2\}$ and $\bn_{\usu}^3(u^*)=\{s_3\}$; when handling $d_1$ by the DSD meta-path in Fig. \ref{fig:1}(b), we can obtain $\bn_{\dsd}^1(d_1)=\{s_1,s_2\}$ and $\bn_{\dsd}^2(d_1)=\{d_2,d_3\}$. In practice, to reduce computational burden, we often restrict the maximum number of neighbors in meta-path guided sampling, i.e., uniformly sampling from the neighbors if the neighbor number exceeds the threshold. For example, if we set the maximum neighbor number for each node as $5$ in USU, then $\max\{|\bn_{\usu}^3(u)|\} = 5 \times 5 \times 5 = 125$. As we consider high order interactions by meta-path, it realizes reasoning by involving the rich structural information in the graph. In the sequel, we will discuss how to design the message construction and passing functions for it.

\subsection{Embedding Propagation \& Aggregation}
We next elaborate on the embedding propagation mechanism in our framework, which encompasses two main components: \emph{message construction} and \emph{message passing}. Then, we extend this technique to high order embedding propagation.

Fig. \ref{fig:2} illustrates the overall flowchart of the embedding propagation process, through both the USU and DSD meta-paths. It can be identified that GCN follows a layer-wise propagation manner, with layer 0 from the current node and gradually increasing along the meta-path. Taking the disease-symptom (DS) graph as an example, in each layer, new embedding of a node $v$ is established on messages from its one-hop neighbors $v^\prime \in \bn(v)$, as well as a projection from its embedding. Specifically, this process on a layer $l$ can be written as
\begin{equation}
\q_v^{l} = \phi \left( \m_{v\rto v}^l + \frac1{|\bn(v)|}\sum_{v^\prime \in \bn(v)} \m_{v\rto v^\prime}^l \right),
\label{eq:msgpass}
\end{equation}
where $\phi(\cdot)$ is an activation function, e.g., tanh, and the message construction of $m_{v\rto v}$ and $m_{v \rto v^\prime}$ are
\begin{equation}
\begin{split}
\m_{v\rto v}^l  &= \W_1^l \q_v, \\
\m_{v\rto v^\prime}^l &= \W_1^l \q_{v^\prime}^{l+1} + \W_2^l (\q_{v} \odot \q_{v^\prime}^{l+1}) .
\end{split}
\label{eq:msgbuild}
\end{equation}
Here, $\W_1$ is responsible for projecting 
$\q_v$ and $\q_{v^\prime}$ into the same space, and the element-wise product $\odot$ measures similarity between the $\q_v$ and $\q_{v^\prime}$. In the DS graph, the propagation follows the rule mentioned in Eqs. \eqref{eq:msgpass} and \eqref{eq:msgbuild}, as the messages involve $\m_{s\rto d}$, $\m_{d\rto s}$, $\m_{d\rto d}$ and $\m_{s\rto s}$. Specifically, the initial embedding of disease is $\q_d^L := \q_d$, which comes from the parametric embedding matrix $\mathcal{Q}$. 

The message propagation is slightly different in the user-symptom (US) graph because we do not have trainable embeddings for users. Instead, user embedding only depends on the message passing and aggregation from symptom nodes. In other words, a user's embedding is generated by the neighboring symptoms. This design allows inductive learning that aims for coping with cold-start users. In particular, when the message is targeted to users, $\q_u^{l-1}$ is
\begin{equation}
\begin{split}
\q_u^{l-1} = \frac1{|\bn(u)|} \sum_{s \in \bn(u)} \m_{u \rto s}^{l} \\
\text{where} \quad  \m_{u \rto s}^{l} = \W_1^l \q_s.
\end{split}
\end{equation}
Since we do not define the initial embeddings for users, no prior information of them is required during this process. When the propagation goes to the output layer $l=0$, it yields the final embedding of the disease $\q^*_d := \q^0_d$ and the user $\q^*_u := \q^0_u$, which will be utilized for score estimation similar to Eq. \eqref{eq:getscore}, i.e., $\hat{y}_{u,d} = \q^{* \top}_u \q^*_d$.

\begin{figure}[t]
\centering
\includegraphics[width=0.48\textwidth]{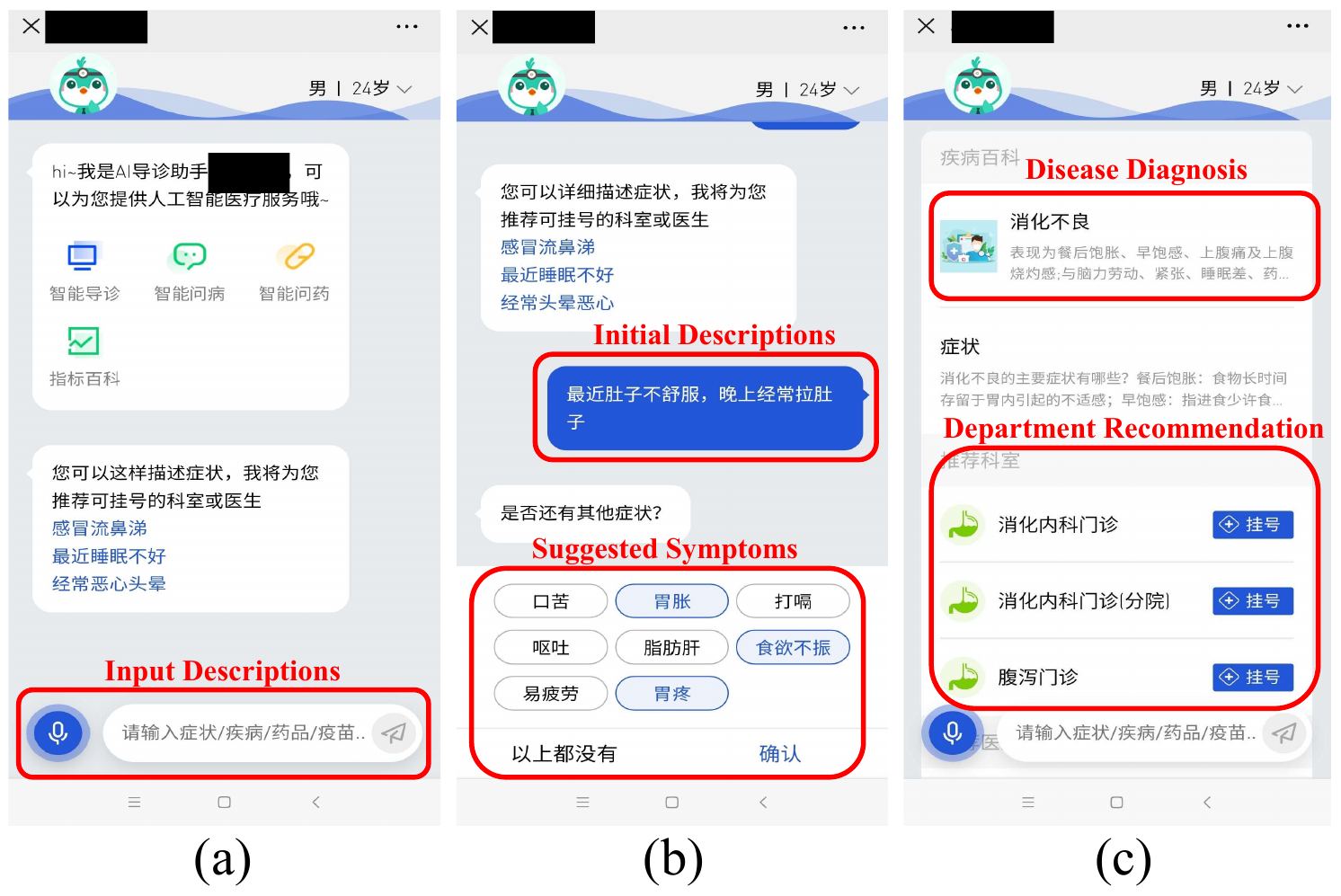}
\caption{The online self-diagnosis system: (a) a user inputs the initial descriptions; (b) the guidance system recommends several alternative symptoms for user to select; (c) after several rounds of Q\&A, the system collects sufficient information and makes a diagnosis. \label{fig:3}}
\end{figure}

\subsection{Optimization}
We adopt the Bayesian Personalized Ranking (BPR) loss \cite{rendle2012bpr} for the model optimization, which is contrastive, such that it encourages the learned embeddings informative for discriminating positive and negative interactions. Specifically, the BPR loss is
\begin{equation}
L_{\text{BPR}} = \sum_{(u,d,d^\prime) \in \Omega} -\log \sigma(\hat{y}_{u,d} - \hat{y}_{u,d^\prime}) + \lambda \|\Theta\|^2_2,
\end{equation}
where $\Omega=\{(u,d,d^\prime) \mid (u,d) \in \Omega^+, (u,d^\prime) \in \Omega^-\}$ represents the pairwise training set; $\Omega^+$ and $\Omega^-$ are the observed and unobserved interaction sets, respectively; $\sigma(\cdot)$ is the sigmoid function, $\sigma: \mathbb{R} \mapsto (0,1)$; $\Theta$ denotes the trainable parameters, i.e., $\Theta = \{\mathcal{Q}, \{ \W^l_1, \W_2^l\}_{l=1}^L\}$; and $\lambda$ is a hyperparameter that controls the imposed $\ell_2$-regularization. During training, we apply message dropout \cite{wang2019neural} to alleviate over fitting. Specifically, we set a probability of $p$ to randomly drop the elements in messages in Eq. \eqref{eq:msgbuild}. 

Moreover, negative sampling is indispensable for optimizing the BPR loss. Since the distribution of disease is highly long-tailed, instead of uniformly sampling negative examples from the entire set of diseases, we conduct an online hard negative example mining similar to \cite{ying2018graph}, i.e., hard examples are dynamically generated at the end of each epoch. To this end, for each positive user-disease pair $(u,d) \in \Omega^+$, we look for a negative example  $(u,d^\prime) \in \Omega^-$ by ranking diseases according to their similarity with respect to the positive disease $d$, and pick the top ranked $d^\prime$ as a hard example. The similarity between diseases here is defined by cosine similarity of embeddings
\begin{equation}
\text{Sim}(d, d^\prime) = \frac{\q_d^{\top}\q_{d^\prime}}{\|\q_d\|_2\|q_{d^\prime}\|_2} .
\end{equation}
 These hard negative examples are more challenging for the model to rank, thus increasing its capability in discriminating diseases at a fine granularity.

\section{Self-Diagnosis System} \label{sec:selfdiag}
We develop an online disease self-diagnosis system, as its front page shown in Fig. \ref{fig:3}. Unlike learning and predicting on EHR data, information offered by users in online system is usually very limited: when engaged in this system, a user often describes his or her feelings in a non-professional and colloquial form. This ambiguous and limited information causes extreme uncertainty for decision making. To deal with it, we first build an NER system for extracting entities pertinent to symptoms from the raw user dialog. Then the extracted symptoms are mapped to \emph{a standard symptom set} which is aligned with symptom nodes in the built HIN. We further build a guiding system to lead users to provide more information. The system displays several symptoms pertinent to the initial description, from which the user picks the most relevant ones. It could be expected that after several rounds of dialog, it collects enough symptoms of the user and is confident to make the final diagnosis.

In this section, we present the implementation of the NER system and then specifically introduce the symptom retrieval component that is responsible for retrieving and displaying alternative symptoms for users, termed as EHR Graph-based Retrieval System (GraphRet). After that, we present the overall flowchart of our self-diagnosis system.

\begin{figure}[t]
\centering
\includegraphics[width=0.48\textwidth]{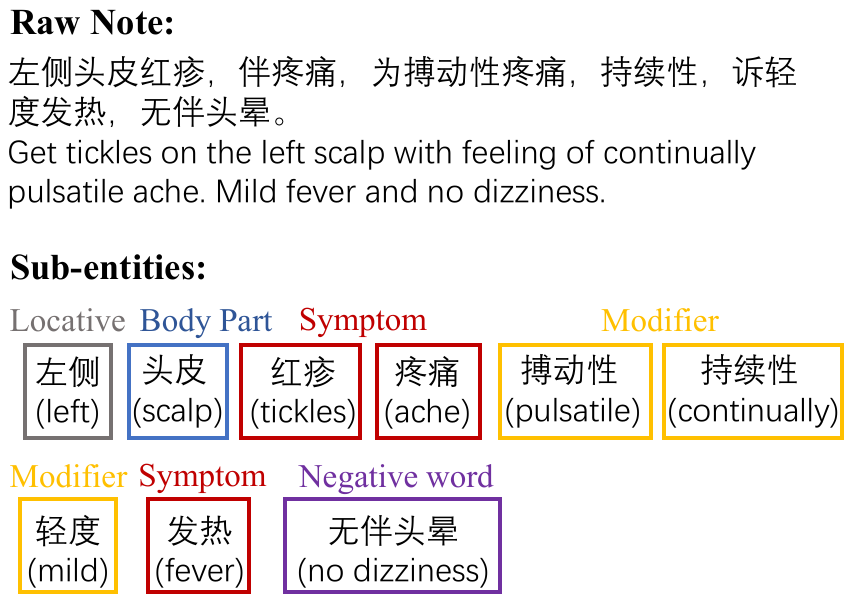}
\caption{An example of the raw clinical note and the extracted entities from it. \label{fig:6}}
\end{figure}

\subsection{Dialog Processing}\label{sec:ner_system}
Different from many general named entity recognition (NER) tasks on public datasets, we have Chinese texts which have many specific medical concepts which rarely appear in common dialogue. What is worse, there has been by far no open datasets about Chinese clinical notes. These problems cause dialog processing in web-based self-diagnosis difficult to be realized based on the current techniques. Accounting for this challenge, We try to collect and label more than 50 thousand medical corpus via crowdsourcing. In particular, in each sentence, 14 types of entities are labeled, including disease, symptom, body part, negative words, etc. An example of labeled sentence is shown in Fig. \ref{fig:6}, where modifier are split from symptom as the independent entities and negative words are specified as a indicator to no such symptom.

Based on this labeled corpus, we train a BiLSTM-CRF  \cite{huang2015bidirectional} model for NER task. In addition, another challenge we identify is that there are multiple extracted symptoms corresponding to similar or same symptom. It commonly appears in self-diagnosis because user tends to input colloquial descriptions, and the semantics could be very diverse. For example, a user may input ``have got a run'' but this phrase is not aligned to any symptom node in the built HIN (the closest phrase in medical terms should be ``diarrhea''). In order to proceed user dialog as accurate as possible, we build a standard symptom set that contains many common equivalent phrases. This BiLSTM-CRF based NER system hence could extract as rich as possible information from the raw dialog, which supports more accurate diagnosis.

\begin{algorithm}[t] 
\caption{Disease Diagnosis with GraphRet and HealGCN. \label{alg:1}}
\begin{algorithmic}[1]
\Require Pretrained graph embeddings $\bq$; The EHR graph $\mathcal{G}$; The GraphRet system $R(\cdot)$;
\State Get embeddings of all diseases $Q_{d}^* \gets \text{ForwardDSD}(\bd;\mathcal{G},\bq)$;
\State Receive the seed symptom set $S_0$ from the user;
\For{epoch $t = 1 \to T$}
\State Retrieve symptoms by GraphRet $\tilde{S} \gets R(S_{t-1}, \mathcal{G})$;
\State  $\hat{S} \gets \text{UserSelect}(\tilde{S})$;
\State Update symptom set $S_t \gets S_{t-1} \cup \hat{S}$;
\State Get the user embedding $\q^*_u \gets \text{ForwardUSU}(S_T;\mathcal{G},\bq)$;
\State Do inference by $\bm{\hat{y}}\gets Q_{d}^*\q_u^{*} \in \mathbb{R}^{|\bd|}$;
\State Exit the loop if the diagnosis confidence is high enough;
\EndFor

\end{algorithmic}
\end{algorithm}

\subsection{Symptom Retrieval}
Symptom retrieval is the bedrock of the diagnosis system for providing alternative symptoms for making final disease diagnosis. We have developed HealGCN where the graph representation learning is tailored towards disease diagnosis. However, the learned embeddings might be suboptimal for symptom retrieval. In this scenario, we would like to perform further representation learning for high quality symptom retrieval, by considering co-occurrence of symptoms in the bipartite disease-symptom (DS) graph. In detail, our task is to generate embeddings of symptoms that can be utilized for nearest-neighbor lookup for top related symptoms. To this end, we apply GraphRet on the DS graph with its symptom and disease embeddings initialized by the pretrained HealGCN. Specifically, we use a max-margin based loss \cite{ying2018graph}
\begin{equation}
L_{\text{MM}} = \mathbb{E}_{s^- \sim P_n(s)}\max \{0, \q_s^{* \top} \q^*_{s^+} - \q_s^{* \top} \q^*_{s^-} + \Delta \},
\end{equation}
where $(s,s^+)$ is a pair of related symptoms; $s^-$ is an unrelated symptom to $s$ sampled from a negative distribution $P_n(s)$; and $\Delta$ is a margin hyperparameter. We keep $\Delta=1.0$ in our experiments. The graph convolutional operation works on the DS graph similar to $\S$ \ref{sec:3} but follows the meta-path symptom-disease-symptom (SDS), to obtain the final symptom embedding $\q_s^*$. Minimizing the max-margin loss encourages the related items to be close, while the unrelated items to be distant in the embedding space. During training, for a sampled symptom $s$, we find the positive symptom $s^+$ that co-occurs most frequently with $s$, while the rest symptoms that co-occur less frequently than $s^+$ are all alternative negative ones. Additionally, we leverage curriculum learning \cite{bengio2009curriculum} to enhance granularity of learned embeddings. In the early training phase, negative symptoms are uniformly sampled, while we gradually involve more hard negative symptoms, i.e., $s^-$ co-occurring with $s$ slightly less than $s^+$, in the subsequent epochs.

\begin{CJK*}{UTF8}{gbsn}
\begin{table*}[t]
  \centering
  \caption{Examples of the clinical notes in collected EHR data. Raw notes are in Chinese, we translate them into English here.}
    \begin{tabular}{|c|p{7em}|p{30em}|p{6.5em}|}
    \hline
    Department & \multicolumn{1}{c|}{Chief Complaint} & \multicolumn{1}{c|}{Clinical Note} & \multicolumn{1}{c|}{Diagnosis Result} \bigstrut\\
    \hline
    \multicolumn{1}{|c|}{\multirow{2}[4]{*}{Gastroenterology}} & 反复咳嗽，腹部不适 & 6月前出现反复咳嗽咳白痰，无发热。接触异味偶有诱发，进食后有咳嗽。咳嗽剧烈伴呕吐，无伴反酸呃逆，间有胸骨后烧灼感，间有空腹时上腹部隐痛，进食隐痛可改善。偶伴鼻塞、打喷嚏、流涕。 & \multicolumn{1}{l|}{胃炎} \bigstrut\\
\cline{2-4}          & Coughing continually and feel stomach uncomfortable. & Got cough six months ago with while phlegm and no fever. Occasionally occurred when touching peculiar smell. Get cough after eating. Sometimes get cough with vomiting, no acid regurgitation and hiccup, feel burning after sternum. Feel dull pain in upper abdomen during fasting, which can be alleviated after eating. Occasionally get nasal congestion, sneezing, and runny nose.  & \multicolumn{1}{l|}{Gastritis} \bigstrut\\
    \hline
    \multicolumn{1}{|c|}{\multirow{2}[4]{*}{Medicine department}} & 反复胸闷、心悸气促伴头晕1周。 & 患者自诉近1周来无诱因反复出现胸闷、心悸不适，伴有头晕及鼻塞，活动后头晕加重，偶有咳嗽咳痰及腹胀不适。 & \multicolumn{1}{l|}{冠心病} \bigstrut\\
\cline{2-4}          & Continually feel chest tightness,palpitation, shortness of breaeth and dizziness for one week. & Got chest tightness, palpitation, dizziness, and nasal congestion with out explicit inducement. Dizziness becomes worse after activity. Occasionally get cough, expectoration, and abdominal distention. & Coronary heart disease \newline{}\newline{} \bigstrut\\
    \hline
    \end{tabular}%
  \label{tab:example_benchmark}%
\end{table*}%
\end{CJK*}

\begin{table}[t]
  \centering
  \caption{Statistics of nodes and edges in our EHR graph.}
    \begin{tabular}{|lr||lr|}
    \hline
    Node & \multicolumn{1}{r||}{Counts} & Edge & \multicolumn{1}{r|}{Counts} \\
    \hline
Disease & 146  & User-Disease & 136,478 \\
        User  & 135,356 & Disease-Symptom & 229,373 \\
       Symptom & 146,871 & User-Symptom & 1,213,475 \\
    \hline
    \end{tabular}%
  \label{tab:datastat}%
\end{table}%

\begin{table}[t]
  \centering
  \caption{Statistics of the most frequent symptoms and diseases in the EHR data. \label{tab:topsympdise}}
\begin{threeparttable}
    \begin{tabular}{|lr||lr|}
    \hline
    Disease & \multicolumn{1}{r||}{Counts} & Symptom & \multicolumn{1}{r|}{Counts} \\
    \hline
    Hypertension    & 12,155      & Fever     & 20,321 \\
    URTI \tnote{*}      & 11,258     & Stomachache     & 11,806 \\
    Pregnancy \tnote{**}     & 9,470     & Vomiting     & 9,173 \\
    Influenza  & 9,083     & Anhelation     & 8,214 \\
    Gastritis   & 6,674     & Runny nose     & 7,031 \\
    Diabetes     & 4,317     & Headache     & 6,937 \\
    Rhinitis  & 2,882     & Coughing     & 6,420 \\
    \hline
    \end{tabular}%
\begin{tablenotes}
\footnotesize
\item[*] Upper Respiratory Tract Infections;
\item[**] Pregnancy is not a disease while it does appear in clinical diagnosis as a health event, we list it here as a ``disease'' for convenience;
\end{tablenotes}
\end{threeparttable}
\end{table}%

Suppose we extract a symptom $s$ from the a Named Entity Recognition (NER) system from the user's description, which is called a \emph{seed symptom}. And the objective of symptom retrieval system is to search for the \emph{related symptoms}. In this work, we propose a hierarchical lookup framework based on the generated symptom embedding $\q_s^*$ through the DS graph. For example, we can take related symptoms from the two-hop neighbors of the seed symptom $s_0$ through the SDS meta-path, i.e., the $\bn_{\text{SDS}}^2(s_0)$, where we define disease $d \in \bn^1_{\text{SDS}}(s_0)$ as the \emph{intermediate node}. This formulation takes the co-occurrence between symptom and disease into account, and allows us to reallocate attention over symptoms conditioned on the intermediate diseases. Technically, considering that there are $m$ diseases linked to the seed symptom $s_0$, we can compute the normalized point-wise mutual information (nPMI) \cite{bouma2009normalized} between symptom $s$ and the connected diseases by
\begin{equation}
h_i := \npmi(s_0;d_i) = \frac{\log({p(s_0, d_i)}/{p(s_0)p(d_i)})}{-\log p(s_0, d_i)},
\end{equation}
where the mutual information $h_i$ denotes the importance of disease $d_i$ to $s_0$. After that, we transform the score into probability distribution
\begin{equation}
p_i = \text{softmax}(h_i) = \frac{e^{h_i}}{\sum_j e^{h_j}} \in [0,1].
\end{equation}
For example, if the target number of retrieved symptoms is $k$, we will pick $k \times p_i$ related symptoms from $\bn(d_i)$, the neighborhood of $d_i$. After that, the related symptoms can be ranked with respect to their cosine similarity to the seed symptom with high efficiency.

\begin{table*}[t]
  \centering
  \caption{Disease diagnosis accuracy of MF \cite{rendle2012bpr}, NeuMF \cite{he2017neural}, GBDT \cite{ke2017lightgbm}, TextCNN \cite{kim2014convolutional}, Med2Vec \cite{choi2016multi}, GraphSAGE \cite{hamilton2017inductive} and our HealGCN, in an offline experiment, where the best ones are in bold.}
\setlength{\tabcolsep}{3mm}{
    \begin{tabular}{lrrrrrrr}
\toprule
          & \multicolumn{1}{l}{Precision@1} & \multicolumn{1}{l}{Recall@3} & \multicolumn{1}{l}{nDCG@3} & \multicolumn{1}{l}{Recall@5} & \multicolumn{1}{l}{nDCG@5} & \multicolumn{1}{l}{Recall@10} & \multicolumn{1}{l}{nDCG@10} \\
\midrule
    MF    & 0.4459 & 0.6816 & 0.5836 & 0.7733 & 0.6214 & 0.8630 & 0.6506 \\
    NeuMF & 0.5089 & 0.7213 & 0.6334 & 0.8019 & 0.6668 & 0.8773 & 0.6913 \\
    GBDT & 0.4990 & 0.7054 & 0.6204 & 0.7657 & 0.6454 & 0.8368 & 0.6683 \\
    TextCNN & 0.5291 & 0.7333 & 0.6491 & 0.8026 & 0.6778 & 0.8734 & 0.7009 \\
    Med2Vec & 0.5218 & 0.7346 & 0.6464 & 0.8103 & 0.6777 & 0.8779 & 0.6999 \\
	GraphSAGE  & 0.5228 & 0.7393 & 0.6504 & 0.8133 & 0.6809 & 0.8872 & 0.7051 \\
    HealGCN & \textBF{0.5507} & \textBF{0.7620} & \textBF{0.6750} & \textBF{0.8339} & \textBF{0.7046} & \textBF{0.9002} & \textBF{0.7263} \\
\bottomrule
    \end{tabular}%
  \label{tab:1}%
}
\vskip -0.1in
\end{table*}%

\begin{table}[t]
  \centering
  \caption{Results of the ablation study.}
    \begin{tabular}{lrrr}
\toprule
          & \multicolumn{1}{l}{Precision@1} & \multicolumn{1}{l}{Recall@3} & \multicolumn{1}{l}{nDCG@3} \\
\midrule
    HealGCN-local & 0.5040 & 0.7263 & 0.6399 \\
    HealGCN-DSD & 0.5225 & 0.7479 & 0.6552 \\
    HealGCN-USU & 0.5495 & 0.7595 & 0.6725 \\
    HealGCN & \textBF{0.5507} & \textBF{0.7620} & \textBF{0.6750} \\
\bottomrule
    \end{tabular}%
  \label{tab:ablation}%
\vskip -0.2in
\end{table}%

\subsection{System Overview}
Fig. \ref{fig:4} illustrates the flowchart of the disease diagnosis system. When a user inputs the initial descriptions, the system turns to request GraphRet for relevant symptoms. After that, the seed symptoms accompanied with the clicked symptoms are adopted for inference by HealGCN, where we generate the user embedding through USU and generate all disease embeddings through DSD. The system then checks if the predicted score of $(u,d)$ calculated by Eq. \eqref{eq:getscore} exceeds the confidence threshold. Predicted probability of each disease can be obtained by a softmax function. If not, the system turns to request GraphRet for more symptoms. It usually happens when users only offer a limited number of symptoms, or their descriptions are ambiguous. After several rounds of Q\&A, the system expands the symptom set and becomes more confident to make the final diagnosis. Our online inference system empowered by the symptom retrieval works as Algorithm \ref{alg:1} shows.

\begin{table*}[t]
  \centering
  \caption{Examples of how retrieved symptoms lead to final diagnosis. The symptoms picked by users are in bold.}
    \begin{tabular}{|l|p{20.46em}|l|l|}
    \hline
    \multicolumn{1}{|l}{\textBF{PMI} \cite{bouma2009normalized}} & \multicolumn{1}{r}{} & \multicolumn{1}{r}{} &  \\
    \hline
    Seed Symptom & \multicolumn{1}{l|}{{Retrieved Symptoms}} & {Top-1 Prediction} & {Groundtruth} \\
    \hline
 Belching    &  \textBF{Stomachache}; Nausea; Regular bowel movement; Weight loss; Vomiting & Gastrointestinal dysfunction
 &  Gastroenteritis \\
\hline
    Chest pain & Muscleache; \textBF{Chest tightness}; Cold; Tinnitus; Hypertension;  &  Coronary artery disease
 & Pulmonary bronchitis
 \\
\hline
Weight loss & \textBF{Constipation}; Abdominal bloating; High blood pressure; Abdomen ache; Navel ache &  Irritable bowel syndrome & Hyperthyroidism \\
\hline
Irritability & \textBF{Insomnia}; \textBF{Tension}; Mental exhaustion; \textBF{Depressed mood}; Shivering & Depressive disorder & Schizophrenia \\
\hline
    \hline
    \multicolumn{1}{|l}{\textBF{GraphRet}} & \multicolumn{1}{r}{} & \multicolumn{1}{r}{} &  \\
    \hline
    {Seed Symptom} & \multicolumn{1}{l|}{{Retrieved Symptoms}} & {Top-1 Prediction} & {Groundtruth} \\
    \hline
    Belching & \textBF{Diarrhea}; \textBF{Dark stool}; \textBF{Bloody stool}; Anorexia; Hematemesis & Gastroenteritis &  Gastroenteritis \\
\hline
    Chest pain & \textBF{Chest tightness}; \textBF{Coughing}; Palpitation; \textBF{Dyspnoea}; Dizziness & Pulmonary bronchitis & Pulmonary bronchitis  \\
\hline
    Weight loss & \textBF{Fatigue}; \textBF{Dry mouth}; Palpitation; \textBF{Impatience}; \textBF{Strong appetite} & Hyperthyroidism & Hyperthyroidism\\
\hline
Irritability & \textBF{Depressed mood}; Alcoholism; \textBF{Phobia}; Hand tremor; \textBF{Fidget} & Schizophrenia & Schizophrenia\\
    \hline
    \end{tabular}%
  \label{tab:case}%
\end{table*}%

\section{Experiments}
In this section, we evaluate HealGCN on the real-world EHR data collected from hospitals offline. Then, we compare our method with several baselines. Specifically, we aim at answering five research questions below:
\begin{itemize}[leftmargin=*, itemsep=0pt, labelsep=5pt]
\item \textbf{RQ1:} Does HealGCN lead to improvement in accuracy of disease diagnosis?
\item \textbf{RQ2:} Does high order neighborhood of HealGCN contribute to better performance?
\item \textbf{RQ3:} How much do meta-paths USU and DSD benefit the inference?
\item \textbf{RQ4:} Does GraphRet lead to good retrieval results?
\item \textbf{RQ5:} How does HealGCN+GraphRet perform in online A/B test?
\end{itemize}

\subsection{Datasets}
We collect many electronic clinical notes from several hospitals and build  the EHR data used in the offline experiments. In each case, there are department name, chief complaint, clinical note written by clinicians, and diagnosis results by clinicians. Examples of this dataset are shown in Table \ref{tab:example_benchmark}. Raw notes are written in Chinese, we here translate them into English for reading. In detail, from each note, we extract the basic personal characteristics of patients, like gender and age. The symptom nodes constitute of entities extracted by NER system mentioned in $\S$ \ref{sec:ner_system}. In this dataset, most patients have only one visit, which exaggerates the diagnosis difficulty substantially. Besides, each note is associated with a disease diagnosis made by physicians, with the corresponding symptom description by patients. We view the diagnosis made by physicians as the groundtruth for learning. The statistics of the EHR graph are shown in Table \ref{tab:datastat} and Table \ref{tab:topsympdise}.

\subsection{Experimental Protocol}
\subsubsection{Evaluation Metric}
We follow the standard evaluation approach in recommendation. During the test phase, the model needs to predict the probability for a user to be affected by each disease, then only those diseases that user really has are taken as positive examples. We adopt widely-accepted metric Recall@k, nDCG@k and P@1, in terms of the top-k ranking results based on the predicted scores, to evaluate the model's performance. 

\subsubsection{Baseline}
As aforementioned, the personal characteristics as well as historical visits are inaccessible in online self-diagnosis task, hence previous transductive learning disease diagnosis methods are \textbf{NOT} applicable, especially those RNN-like models that depend on sequential EHR data, e.g., Doctor AI \cite{choi2016doctor}, RETAIN \cite{choi2016retain}, Dipole \cite{ma2017dipole}, etc. Therefore, we only pick baselines which are inductive or easy to be adapted to the inductive learning manner. Considering this, we compare our HealGCN with the following baselines:

\textbf{MF} \cite{rendle2012bpr} : This is a classical collaborative filtering method that tries to project the discrete index of user and item into the same real-valued vector space, then measures the similarity of user and item embeddings for predicting scores. For the EHR data, we only model the symptom and disease embeddings, and the user's embedding is aggregated by symptoms. The BPR loss is adopted for its optimization.

\textbf{NeuMF} \cite{he2017neural}: It is a neural collaborative filtering model, which uses hidden layers above the user and item embeddings, in order to mine the nonlinear feature interactions. This method, as well as the MF above, serves as representative traditional collaborative filtering approaches for comparison.

\textbf{GBDT} \cite{ke2017lightgbm}: The gradient boosting decision tree (GBDT) model is popular in many industrial applications. We use Word2Vec \cite{mikolov2013distributed} to learn the word embeddings of symptoms, then take average of each user's symptom embeddings as the input of the GBDT model.

\textbf{TextCNN} \cite{kim2014convolutional}: We formulate disease diagnosis as a text classification task based on TextCNN. Specifically, we model the symptom descriptions of a user as ``word'' embeddings, followed by multiple convolutional layers and dense layers targeted to text classification. We include it to compare our method with the NLP-based disease diagnosis approaches.

\textbf{Med2Vec} \cite{choi2016multi}: It is a medical embedding method that learns embeddings of medical codes and visits based on a skip-gram model similar to Word2Vec \cite{mikolov2013distributed}. Different from the original Med2Vec, we here train it on the clinical notes to get the code (symptom) representation. With the pretrained symptom embeddings, we add a multi-layer perceptron to get the final prediction on disease.

\textbf{GraphSAGE} \cite{hamilton2017inductive}: It is an inductive GCN but ignores the heterogeneity of the EHR graph. We perform it for comparing our method with homogeneous GCN in terms of disease diagnosis.

\textbf{HealGCN-local}: This is a reduced version of our HealGCN that only considers the one-hop neighbors in embedding propagation with the EHR graph. We aim to find out how the high order neighbors contribute to the final results by comparing it with HealGCN.

\textbf{HealGCN-USU \& DSD}: Similar to HealGCN-local, these two models are reduced from HealGCN by only involving the USU or DSD meta-path. They are used to measuring the contribution from different meta-paths.

\subsubsection{Hyperparameters}
We implement all models on PyTorch \cite{paszke2019pytorch}. The size of symptom and disease embeddings is fixed at $64$. The Adam optimizer \cite{kingma2014adam} is used for optimization of all methods. We apply a grid search for optimal hyperparameters: learning rate in $\{0.05, 0.01, 0.005,0.001\}$, batch size in $\{128,256,512,1024,2048\}$, and weight decay in $\{10^{-5},10^{-4},10^{-3}\}$. In terms of the HealGCN model, the maximum numbers of neighbors in the USU and DSD meta-paths are $\{5,5,5\}$ and $\{20,5\}$, respectively. We split the full data into the training, validation and test sets by $7:1:2$. During data processing, we only involve the symptoms appearing in the training set for preventing data leakage. During training, we evaluate the model's nDCG@5 on the validation set for early stopping.

\subsection{RQ1: Overall Comparison} \label{sec:overallexp}
In this section, we compare the overall performance of our HealGCN and the selected baselines, as shown in Table \ref{tab:1}. In this experiment, all symptoms extracted from the clinical notes are used for disease diagnosis, without retrieved symptoms from GraphRet. It can be observed that our HealGCN consistently outperforms other baselines, with improvement around $5\%$ over the best baselines with respect to all metrics. MF only takes direct multiplication of user and disease embeddings, while NeuMF performs better by  leveraging hidden layers and exploiting high order interactions between embeddings. This demonstrates the effectiveness of non-linear interactions between user and disease embeddings. TextCNN performs relatively well, but does not involve connectivity between user and disease considering the graphical structure. Med2Vec takes semantics of symptoms into consideration by utilizing a Word2Vec-like model; however, it ignores rich interactions in the EHR graph and does not perform very well. GraphSAGE leverages graph structure, but ignores the interaction types and performs worse than our method.

\subsection{RQ2 \& 3: Ablation Study}
We perform ablation experiments to explore influences of different components of our HealGCN, with results shown in Table \ref{tab:ablation}. Recall that the local model only involves the one-hop neighbors in message passing and aggregation, it does not exploit high order neighbors as well as the node's position in the graph. Therefore, the local model performs the worst among all.

\begin{table}[t]
  \centering
  \caption{Rec@20, Rec@50 and Rec@100 achieved by GraphRet and compared baselines on symptom retrieval.}
    \begin{tabular}{lrrr}
\toprule
          & \multicolumn{1}{l}{Recall@20} & \multicolumn{1}{l}{Recall@50} & \multicolumn{1}{l}{Recall@100} \\
\midrule
    PMI \cite{bouma2009normalized} & 0.0786 & 0.1063 & 0.1298 \\
		Med2Vec \cite{choi2016multi} & 0.1253 &	0.1799 &	0.2332 \\
    GraphRet (ours) &	\textBF{0.2798} & \textBF{0.3953} & \textBF{0.4855} \\
\bottomrule
    \end{tabular}%
  \label{tab:retrecall}%
\end{table}%

\begin{table}[t]
  \centering
  \caption{P@1, Rec@5 and nDCG@5 achieved by HealGCN supported by GraphRet, compared with the Direct method.}
    \begin{tabular}{lrrr}
\toprule
          & \multicolumn{1}{l}{Precision@1} & \multicolumn{1}{l}{Recall@5} & \multicolumn{1}{l}{nDCG@5} \\
\midrule
    Direct & 0.1623 & 0.3794 & 0.2729 \\
    GraphRet &	\textBF{0.2719} & \textBF{0.5230} & \textBF{0.4053} \\
\bottomrule
    \end{tabular}%
  \label{tab:resofret}%
\end{table}%

Furthermore, both DSD and USU models perform worse than full HealGCN in general, and better than the local model. In particular, we identify that the USU model is much better than the DSD model, which indicates the importance of taking symptom co-occurrence through USU into account to encode users into better embeddings. In summary, by combining both DSD and USU meta-paths, our HealGCN achieves the best result.

\subsection{RQ4: Retrieval System}
We compare our GraphRet with Med2Vec and PMI in a simulated online test. In Med2Vec and PMI, we directly compute cosine similarity and normalized PMI (npmi) \cite{bouma2009normalized} between all symptoms and the initial one, respectively. We evaluate the retrieval accuracy by counting how many retrieved symptoms are aligned with the true symptoms of users, as shwon in Table \ref{tab:retrecall}. It can be identified that GraphRet significantly outperforms the baselines. We list several cases in Table \ref{tab:case}. It can be observed that users pick more retrieved symptoms from GraphRet, thus the expanded symptom set leads to accurate prediction by HealGCN. Besides, GraphRet can yield more diverse symptoms than PMI, which benefits in reducing diagnosis mistakes.

Moreover, we evaluate how much the retrieved symptoms benefit in disease diagnosis accuracy, results are illustrated in Table \ref{tab:resofret}. Note that different from the experiment done in $\S$ \ref{sec:overallexp}, here we use only one symptom as the seed symptom, and the Direct method performs inference only on the seed symptom. In GraphRet, inference is performed involving all retrieved symptoms. It can be identified that retrieval systems significantly benefit the inference accuracy, which verifies the effectiveness of retrieval system for improving diagnosis accuracy.

\subsection{RQ5: Online A/B Test}
We evaluate the proposed HealGCN+GraphRet framework on a real-world online self-diagnosis service platform, which currently serves around 60 thousand requests per day, in an online A/B test for one month. On this platform, users are made diagnosis and then guided to make an appointment with a right physician in a right department, which covers over 500 hospitals right now. The base model that has already developed for demo on this platform constitute of TextCNN for disease diagnosis and PMI for symptom retrieval. 

In the A/B test, we split requests into two buckets: one for the base model and other for the proposed method. Several indicators are adopted: user-physician (UP) click ratio, user-disease (UD) click ratio and user-order (UO) click ratio, because our system displays not only disease diagnosis, but also department and physician recommendations. UP and UD means how many users click the recommended physicians and diseases for detailed exploration, respectively. UO means how many users make appointment to the recommended departments for further consultation. These indicators are useful proxy for measuring diagnosis accuracy. Moreover, we send questionaire to users for feedback about their diagnosis results after they come to see a doctor, then we obtain an accuracy of the diagnosis results, namely ACC, from the feedback. It can be observed from Table \ref{tab:abtest} that the proposed method outperforms the base model significantly, which shows our model has advantage in disease diagnosis, thus attracting users to make further exploration. And our new system indeed obtains better accuracy in practice.

\begin{table}[t]
  \centering
  \caption{Online evaluation results of the proposed method with the existing base model (TextCNN+PMI). Here, UP, UD and UO denote user click ratio regarding the recommended physician, disease and appointment order, respectively. ACC indicates accuracy of those who send back feedback after come to see a doctor.}
    \begin{tabular}{lrrrr}
\toprule
          & \multicolumn{1}{l}{UP} & \multicolumn{1}{l}{UD} & \multicolumn{1}{l}{UO} & \multicolumn{1}{l}{ACC}\\
\midrule
    Base Model& 74.4\% & 20.6\% & 4.8\% & 35.5\% \\
    HealGCN+GraphRet &	86.6\% & 26.8\% & 6.9\% & 42.3\% \\
\bottomrule
    \end{tabular}%
  \label{tab:abtest}%
\end{table}%

\section{Conclusion}
In this paper, we proposed HealGCN for disease diagnosis and GraphRet for symptom retrieval aiming at handling two main challenges in online self-diagnosis: cold-start users and ambiguous descriptions. HealGCN adopts an inductive learning paradigm, thus applicable for users without historical hospital visits. GraphRet serves for symptom retrieval in Q\&A in self-diagnosis by applying graph convolution operations to a bipartite disease-symptom graph to generate symptom embeddings for retrieval. The offline evaluation and online test verified the superiority of HealGCN in diagnosis accuracy. It also proved that GraphRet benefits in providing a rich supportive symptom set for users to select, resulting in more accurate diagnosis.
\section*{Acknowledgment}
The research of Shao-Lun Huang was supported in part by the Natural Science Foundation of China under Grant 61807021, in part by the Shenzhen Science and Technology Program under Grant KQTD20170810150821146, and in part by the Innovation and Entrepreneurship Project for Overseas High-Level Talents of Shenzhen under Grant KQJSCX20180327144037831.

\bibliographystyle{ACM-Reference-Format}
\bibliography{sample-base}


\begin{thebibliography}{45}


\ifx \showCODEN    \undefined \def \showCODEN     #1{\unskip}     \fi
\ifx \showDOI      \undefined \def \showDOI       #1{#1}\fi
\ifx \showISBNx    \undefined \def \showISBNx     #1{\unskip}     \fi
\ifx \showISBNxiii \undefined \def \showISBNxiii  #1{\unskip}     \fi
\ifx \showISSN     \undefined \def \showISSN      #1{\unskip}     \fi
\ifx \showLCCN     \undefined \def \showLCCN      #1{\unskip}     \fi
\ifx \shownote     \undefined \def \shownote      #1{#1}          \fi
\ifx \showarticletitle \undefined \def \showarticletitle #1{#1}   \fi
\ifx \showURL      \undefined \def \showURL       {\relax}        \fi
\providecommand\bibfield[2]{#2}
\providecommand\bibinfo[2]{#2}
\providecommand\natexlab[1]{#1}
\providecommand\showeprint[2][]{arXiv:#2}

\bibitem[\protect\citeauthoryear{Bengio, Louradour, Collobert, and
  Weston}{Bengio et~al\mbox{.}}{2009}]%
        {bengio2009curriculum}
\bibfield{author}{\bibinfo{person}{Yoshua Bengio},
  \bibinfo{person}{J{\'e}r{\^o}me Louradour}, \bibinfo{person}{Ronan
  Collobert}, {and} \bibinfo{person}{Jason Weston}.}
  \bibinfo{year}{2009}\natexlab{}.
\newblock \showarticletitle{Curriculum learning}. In
  \bibinfo{booktitle}{\emph{Proceedings of the 26th Annual International
  Conference on Machine Learning}}. \bibinfo{pages}{41--48}.
\newblock


\bibitem[\protect\citeauthoryear{Biswal, Xiao, Glass, Milkovits, and
  Sun}{Biswal et~al\mbox{.}}{2019}]%
        {biswal2019doctor2vec}
\bibfield{author}{\bibinfo{person}{Siddharth Biswal}, \bibinfo{person}{Cao
  Xiao}, \bibinfo{person}{Lucas~M. Glass}, \bibinfo{person}{Elizabeth
  Milkovits}, {and} \bibinfo{person}{Jimeng Sun}.}
  \bibinfo{year}{2019}\natexlab{}.
\newblock \showarticletitle{Doctor2Vec: Dynamic doctor representation learning
  for clinical trial recruitment}.
\newblock \bibinfo{journal}{\emph{arXiv preprint arXiv:1911.10395}}
  (\bibinfo{year}{2019}).
\newblock


\bibitem[\protect\citeauthoryear{Bouma}{Bouma}{2009}]%
        {bouma2009normalized}
\bibfield{author}{\bibinfo{person}{Gerlof Bouma}.}
  \bibinfo{year}{2009}\natexlab{}.
\newblock \showarticletitle{Normalized (pointwise) mutual information in
  collocation extraction}.
\newblock \bibinfo{journal}{\emph{Proceedings of German Society for
  Computational Linguistics and Language Technology}}, \bibinfo{pages}{31--40}.
\newblock


\bibitem[\protect\citeauthoryear{Che and Liu}{Che and Liu}{2017}]%
        {che2017deep}
\bibfield{author}{\bibinfo{person}{Zhengping Che} {and} \bibinfo{person}{Yan
  Liu}.} \bibinfo{year}{2017}\natexlab{}.
\newblock \showarticletitle{Deep learning solutions to computational
  phenotyping in health care}. In \bibinfo{booktitle}{\emph{IEEE International
  Conference on Data Mining Workshops}}. \bibinfo{pages}{1100--1109}.
\newblock


\bibitem[\protect\citeauthoryear{Choi, Bahadori, Schuetz, Stewart, and
  Sun}{Choi et~al\mbox{.}}{2016a}]%
        {choi2016doctor}
\bibfield{author}{\bibinfo{person}{Edward Choi}, \bibinfo{person}{Mohammad~Taha
  Bahadori}, \bibinfo{person}{Andy Schuetz}, \bibinfo{person}{Walter~F
  Stewart}, {and} \bibinfo{person}{Jimeng Sun}.}
  \bibinfo{year}{2016}\natexlab{a}.
\newblock \showarticletitle{Doctor {AI}: Predicting clinical events via
  recurrent neural networks}. In \bibinfo{booktitle}{\emph{Machine Learning for
  Healthcare Conference}}. \bibinfo{pages}{301--318}.
\newblock


\bibitem[\protect\citeauthoryear{Choi, Bahadori, Searles, Coffey, Thompson,
  Bost, Tejedor-Sojo, and Sun}{Choi et~al\mbox{.}}{2016b}]%
        {choi2016multi}
\bibfield{author}{\bibinfo{person}{Edward Choi}, \bibinfo{person}{Mohammad~Taha
  Bahadori}, \bibinfo{person}{Elizabeth Searles}, \bibinfo{person}{Catherine
  Coffey}, \bibinfo{person}{Michael Thompson}, \bibinfo{person}{James Bost},
  \bibinfo{person}{Javier Tejedor-Sojo}, {and} \bibinfo{person}{Jimeng Sun}.}
  \bibinfo{year}{2016}\natexlab{b}.
\newblock \showarticletitle{Multi-layer representation learning for medical
  concepts}. In \bibinfo{booktitle}{\emph{Proceedings of the 22nd ACM SIGKDD
  International Conference on Knowledge Discovery and Data Mining}}.
  \bibinfo{pages}{1495--1504}.
\newblock


\bibitem[\protect\citeauthoryear{Choi, Bahadori, Sun, Kulas, Schuetz, and
  Stewart}{Choi et~al\mbox{.}}{2016c}]%
        {choi2016retain}
\bibfield{author}{\bibinfo{person}{Edward Choi}, \bibinfo{person}{Mohammad~Taha
  Bahadori}, \bibinfo{person}{Jimeng Sun}, \bibinfo{person}{Joshua Kulas},
  \bibinfo{person}{Andy Schuetz}, {and} \bibinfo{person}{Walter Stewart}.}
  \bibinfo{year}{2016}\natexlab{c}.
\newblock \showarticletitle{{RETAIN}: An interpretable predictive model for
  healthcare using reverse time attention mechanism}. In
  \bibinfo{booktitle}{\emph{Advances in Neural Information Processing
  Systems}}. \bibinfo{pages}{3504--3512}.
\newblock


\bibitem[\protect\citeauthoryear{Choi, Xiao, Stewart, and Sun}{Choi
  et~al\mbox{.}}{2018}]%
        {choi2018mime}
\bibfield{author}{\bibinfo{person}{Edward Choi}, \bibinfo{person}{Cao Xiao},
  \bibinfo{person}{Walter Stewart}, {and} \bibinfo{person}{Jimeng Sun}.}
  \bibinfo{year}{2018}\natexlab{}.
\newblock \showarticletitle{{MiME}: Multilevel medical embedding of electronic
  health records for predictive healthcare}. In
  \bibinfo{booktitle}{\emph{Advances in Neural Information Processing
  Systems}}. \bibinfo{pages}{4547--4557}.
\newblock


\bibitem[\protect\citeauthoryear{Choi, Xu, Li, Dusenberry, Flores, Xue, and
  Dai}{Choi et~al\mbox{.}}{2019}]%
        {choi2019graph}
\bibfield{author}{\bibinfo{person}{Edward Choi}, \bibinfo{person}{Zhen Xu},
  \bibinfo{person}{Yujia Li}, \bibinfo{person}{Michael~W. Dusenberry},
  \bibinfo{person}{Gerardo Flores}, \bibinfo{person}{Yuan Xue}, {and}
  \bibinfo{person}{Andrew~M. Dai}.} \bibinfo{year}{2019}\natexlab{}.
\newblock \showarticletitle{Graph convolutional transformer: Learning the
  graphical structure of electronic health records}.
\newblock \bibinfo{journal}{\emph{arXiv preprint arXiv:1906.04716}}
  (\bibinfo{year}{2019}).
\newblock


\bibitem[\protect\citeauthoryear{Choi, Chiu, and Sontag}{Choi
  et~al\mbox{.}}{2016d}]%
        {choi2016learning}
\bibfield{author}{\bibinfo{person}{Youngduck Choi}, \bibinfo{person}{Chill Yi-I
  Chiu}, {and} \bibinfo{person}{David Sontag}.}
  \bibinfo{year}{2016}\natexlab{d}.
\newblock \showarticletitle{Learning low-dimensional representations of medical
  concepts}.
\newblock \bibinfo{journal}{\emph{AMIA Summits on Translational Science
  Proceedings}} (\bibinfo{year}{2016}), \bibinfo{pages}{41--50}.
\newblock


\bibitem[\protect\citeauthoryear{Devlin, Chang, Lee, and Toutanova}{Devlin
  et~al\mbox{.}}{2019}]%
        {devlin2018bert}
\bibfield{author}{\bibinfo{person}{Jacob Devlin}, \bibinfo{person}{Ming-Wei
  Chang}, \bibinfo{person}{Kenton Lee}, {and} \bibinfo{person}{Kristina
  Toutanova}.} \bibinfo{year}{2019}\natexlab{}.
\newblock \showarticletitle{{BERT}: Pre-training of deep bidirectional
  transformers for language understanding}. In
  \bibinfo{booktitle}{\emph{Proceedings of the Conference of the Association
  for Computational Linguistics}}. \bibinfo{pages}{4171--4186}.
\newblock


\bibitem[\protect\citeauthoryear{Dligach and Miller}{Dligach and
  Miller}{2018}]%
        {dligach2018learning}
\bibfield{author}{\bibinfo{person}{Dmitriy Dligach} {and}
  \bibinfo{person}{Timothy Miller}.} \bibinfo{year}{2018}\natexlab{}.
\newblock \showarticletitle{Learning patient representations from text}.
\newblock \bibinfo{journal}{\emph{arXiv preprint arXiv:1805.02096}}
  (\bibinfo{year}{2018}).
\newblock


\bibitem[\protect\citeauthoryear{Farhan, Wang, Huang, Wang, Wang, and
  Jiang}{Farhan et~al\mbox{.}}{2016}]%
        {farhan2016predictive}
\bibfield{author}{\bibinfo{person}{Wael Farhan}, \bibinfo{person}{Zhimu Wang},
  \bibinfo{person}{Yingxiang Huang}, \bibinfo{person}{Shuang Wang},
  \bibinfo{person}{Fei Wang}, {and} \bibinfo{person}{Xiaoqian Jiang}.}
  \bibinfo{year}{2016}\natexlab{}.
\newblock \showarticletitle{A predictive model for medical events based on
  contextual embedding of temporal sequences}.
\newblock \bibinfo{journal}{\emph{JMIR Medical Informatics}}
  \bibinfo{volume}{4}, \bibinfo{number}{4} (\bibinfo{year}{2016}),
  \bibinfo{pages}{e39}.
\newblock


\bibitem[\protect\citeauthoryear{Galk{\'o} and Eickhoff}{Galk{\'o} and
  Eickhoff}{2018}]%
        {galko2018biomedical}
\bibfield{author}{\bibinfo{person}{Ferenc Galk{\'o}} {and}
  \bibinfo{person}{Carsten Eickhoff}.} \bibinfo{year}{2018}\natexlab{}.
\newblock \showarticletitle{Biomedical question answering via weighted neural
  network passage retrieval}. In \bibinfo{booktitle}{\emph{European Conference
  on Information Retrieval}}. \bibinfo{pages}{523--528}.
\newblock


\bibitem[\protect\citeauthoryear{Hamilton, Ying, and Leskovec}{Hamilton
  et~al\mbox{.}}{2017}]%
        {hamilton2017inductive}
\bibfield{author}{\bibinfo{person}{Will Hamilton}, \bibinfo{person}{Zhitao
  Ying}, {and} \bibinfo{person}{Jure Leskovec}.}
  \bibinfo{year}{2017}\natexlab{}.
\newblock \showarticletitle{Inductive representation learning on large graphs}.
  In \bibinfo{booktitle}{\emph{Advances in Neural Information Processing
  Systems}}. \bibinfo{pages}{1024--1034}.
\newblock


\bibitem[\protect\citeauthoryear{Han, Sun, Yan, and Yu}{Han
  et~al\mbox{.}}{2010}]%
        {han2010mining}
\bibfield{author}{\bibinfo{person}{Jiawei Han}, \bibinfo{person}{Yizhou Sun},
  \bibinfo{person}{Xifeng Yan}, {and} \bibinfo{person}{Philip~S. Yu}.}
  \bibinfo{year}{2010}\natexlab{}.
\newblock \showarticletitle{Mining knowledge from databases: An information
  network analysis approach}. In \bibinfo{booktitle}{\emph{Proceedings of the
  ACM SIGMOD International Conference on Management of Data}}.
  \bibinfo{pages}{1251--1252}.
\newblock


\bibitem[\protect\citeauthoryear{He, Liao, Zhang, Nie, Hu, and Chua}{He
  et~al\mbox{.}}{2017}]%
        {he2017neural}
\bibfield{author}{\bibinfo{person}{Xiangnan He}, \bibinfo{person}{Lizi Liao},
  \bibinfo{person}{Hanwang Zhang}, \bibinfo{person}{Liqiang Nie},
  \bibinfo{person}{Xia Hu}, {and} \bibinfo{person}{Tat-Seng Chua}.}
  \bibinfo{year}{2017}\natexlab{}.
\newblock \showarticletitle{Neural collaborative filtering}. In
  \bibinfo{booktitle}{\emph{Proceedings of the 26th International Conference on
  World Wide Web}}. \bibinfo{pages}{173--182}.
\newblock


\bibitem[\protect\citeauthoryear{Herrett, Gallagher, Bhaskaran, Forbes, Mathur,
  van Staa, and Smeeth}{Herrett et~al\mbox{.}}{2015}]%
        {herrett2015data}
\bibfield{author}{\bibinfo{person}{Emily Herrett}, \bibinfo{person}{Arlene~M.
  Gallagher}, \bibinfo{person}{Krishnan Bhaskaran}, \bibinfo{person}{Harriet
  Forbes}, \bibinfo{person}{Rohini Mathur}, \bibinfo{person}{Tjeerd van Staa},
  {and} \bibinfo{person}{Liam Smeeth}.} \bibinfo{year}{2015}\natexlab{}.
\newblock \showarticletitle{Data resource profile: Clinical practice research
  datalink (CPRD)}.
\newblock \bibinfo{journal}{\emph{International Journal of Epidemiology}}
  \bibinfo{volume}{44} (\bibinfo{year}{2015}), \bibinfo{pages}{827--836}.
\newblock


\bibitem[\protect\citeauthoryear{Hettige, Wang, Li, Le, and Buntine}{Hettige
  et~al\mbox{.}}{2020}]%
        {hettige2019medgraph}
\bibfield{author}{\bibinfo{person}{Bhagya Hettige}, \bibinfo{person}{Weiqing
  Wang}, \bibinfo{person}{Yuan-Fang Li}, \bibinfo{person}{Suong Le}, {and}
  \bibinfo{person}{Wray Buntine}.} \bibinfo{year}{2020}\natexlab{}.
\newblock \showarticletitle{{MedGraph}: Structural and Temporal Representation
  Learning of Electronic Medical Records}. In
  \bibinfo{booktitle}{\emph{Proceedings of the 24th European Conference on
  Artificial Intelligence (ECAI)}}.
\newblock


\bibitem[\protect\citeauthoryear{Hosseini, Chen, Wu, Sun, and
  Sarrafzadeh}{Hosseini et~al\mbox{.}}{2018}]%
        {hosseini2018heteromed}
\bibfield{author}{\bibinfo{person}{Anahita Hosseini}, \bibinfo{person}{Ting
  Chen}, \bibinfo{person}{Wenjun Wu}, \bibinfo{person}{Yizhou Sun}, {and}
  \bibinfo{person}{Majid Sarrafzadeh}.} \bibinfo{year}{2018}\natexlab{}.
\newblock \showarticletitle{Hetero{M}ed: Heterogeneous information network for
  medical diagnosis}. In \bibinfo{booktitle}{\emph{Proceedings of the 27th ACM
  International Conference on Information and Knowledge Management}}.
  \bibinfo{pages}{763--772}.
\newblock


\bibitem[\protect\citeauthoryear{Hosseini, Davis, and Sarrafzadeh}{Hosseini
  et~al\mbox{.}}{2019}]%
        {hosseini2019hierarchical}
\bibfield{author}{\bibinfo{person}{Anahita Hosseini}, \bibinfo{person}{Tyler
  Davis}, {and} \bibinfo{person}{Majid Sarrafzadeh}.}
  \bibinfo{year}{2019}\natexlab{}.
\newblock \showarticletitle{Hierarchical target-attentive diagnosis prediction
  in heterogeneous information networks}. In
  \bibinfo{booktitle}{\emph{International Conference on Data Mining
  Workshops}}. \bibinfo{pages}{949--957}.
\newblock


\bibitem[\protect\citeauthoryear{Huang, Xu, and Yu}{Huang
  et~al\mbox{.}}{2015}]%
        {huang2015bidirectional}
\bibfield{author}{\bibinfo{person}{Zhiheng Huang}, \bibinfo{person}{Wei Xu},
  {and} \bibinfo{person}{Kai Yu}.} \bibinfo{year}{2015}\natexlab{}.
\newblock \showarticletitle{Bidirectional LSTM-CRF models for sequence
  tagging}.
\newblock \bibinfo{journal}{\emph{arXiv preprint arXiv:1508.01991}}
  (\bibinfo{year}{2015}).
\newblock


\bibitem[\protect\citeauthoryear{Johnson, Pollard, Shen, Li-wei, Feng,
  Ghassemi, Moody, Szolovits, Celi, and Mark}{Johnson et~al\mbox{.}}{2016}]%
        {johnson2016mimic}
\bibfield{author}{\bibinfo{person}{Alistair~EW. Johnson},
  \bibinfo{person}{Tom~J. Pollard}, \bibinfo{person}{Lu Shen},
  \bibinfo{person}{H.~Lehman Li-wei}, \bibinfo{person}{Mengling Feng},
  \bibinfo{person}{Mohammad Ghassemi}, \bibinfo{person}{Benjamin Moody},
  \bibinfo{person}{Peter Szolovits}, \bibinfo{person}{Leo~Anthony Celi}, {and}
  \bibinfo{person}{Roger~G. Mark}.} \bibinfo{year}{2016}\natexlab{}.
\newblock \showarticletitle{MIMIC-III, a freely accessible critical care
  database}.
\newblock \bibinfo{journal}{\emph{Scientific Data}}  \bibinfo{volume}{3}
  (\bibinfo{year}{2016}).
\newblock


\bibitem[\protect\citeauthoryear{Ke, Meng, Finley, Wang, Chen, Ma, Ye, and
  Liu}{Ke et~al\mbox{.}}{2017}]%
        {ke2017lightgbm}
\bibfield{author}{\bibinfo{person}{Guolin Ke}, \bibinfo{person}{Qi Meng},
  \bibinfo{person}{Thomas Finley}, \bibinfo{person}{Taifeng Wang},
  \bibinfo{person}{Wei Chen}, \bibinfo{person}{Weidong Ma},
  \bibinfo{person}{Qiwei Ye}, {and} \bibinfo{person}{Tie-Yan Liu}.}
  \bibinfo{year}{2017}\natexlab{}.
\newblock \showarticletitle{Lightgbm: A highly efficient gradient boosting
  decision tree}. In \bibinfo{booktitle}{\emph{Advances in Neural Information
  Processing Systems}}. \bibinfo{pages}{3146--3154}.
\newblock


\bibitem[\protect\citeauthoryear{Kim}{Kim}{2014}]%
        {kim2014convolutional}
\bibfield{author}{\bibinfo{person}{Yoon Kim}.} \bibinfo{year}{2014}\natexlab{}.
\newblock \showarticletitle{Convolutional neural networks for sentence
  classification}.
\newblock \bibinfo{journal}{\emph{arXiv preprint arXiv:1408.5882}}
  (\bibinfo{year}{2014}).
\newblock


\bibitem[\protect\citeauthoryear{Kingma and Ba}{Kingma and Ba}{2014}]%
        {kingma2014adam}
\bibfield{author}{\bibinfo{person}{Diederik~P. Kingma} {and}
  \bibinfo{person}{Jimmy Ba}.} \bibinfo{year}{2014}\natexlab{}.
\newblock \showarticletitle{Adam: A method for stochastic optimization}.
\newblock \bibinfo{journal}{\emph{arXiv preprint arXiv:1412.6980}}
  (\bibinfo{year}{2014}).
\newblock


\bibitem[\protect\citeauthoryear{Kipf and Welling}{Kipf and Welling}{2016}]%
        {kipf2016semi}
\bibfield{author}{\bibinfo{person}{Thomas~N. Kipf} {and} \bibinfo{person}{Max
  Welling}.} \bibinfo{year}{2016}\natexlab{}.
\newblock \showarticletitle{Semi-supervised classification with graph
  convolutional networks}.
\newblock \bibinfo{journal}{\emph{arXiv preprint arXiv:1609.02907}}
  (\bibinfo{year}{2016}).
\newblock


\bibitem[\protect\citeauthoryear{Li, Rao, Solares, Hassaine, Canoy, Zhu,
  Rahimi, and Salimi-Khorshidi}{Li et~al\mbox{.}}{2019}]%
        {li2019behrt}
\bibfield{author}{\bibinfo{person}{Yikuan Li}, \bibinfo{person}{Shishir Rao},
  \bibinfo{person}{Jose Roberto~Ayala Solares}, \bibinfo{person}{Abdelaali
  Hassaine}, \bibinfo{person}{Dexter Canoy}, \bibinfo{person}{Yajie Zhu},
  \bibinfo{person}{Kazem Rahimi}, {and} \bibinfo{person}{Gholamreza
  Salimi-Khorshidi}.} \bibinfo{year}{2019}\natexlab{}.
\newblock \showarticletitle{{BEHRT}: Transformer for electronic health
  records}.
\newblock \bibinfo{journal}{\emph{arXiv preprint arXiv:1907.09538}}
  (\bibinfo{year}{2019}).
\newblock


\bibitem[\protect\citeauthoryear{Lin}{Lin}{2009}]%
        {lin2009intelligent}
\bibfield{author}{\bibinfo{person}{Rong-Ho Lin}.}
  \bibinfo{year}{2009}\natexlab{}.
\newblock \showarticletitle{An intelligent model for liver disease diagnosis}.
\newblock \bibinfo{journal}{\emph{Artificial Intelligence in Medicine}}
  \bibinfo{volume}{47} (\bibinfo{year}{2009}), \bibinfo{pages}{53--62}.
\newblock


\bibitem[\protect\citeauthoryear{Ma, Chitta, Zhou, You, Sun, and Gao}{Ma
  et~al\mbox{.}}{2017}]%
        {ma2017dipole}
\bibfield{author}{\bibinfo{person}{Fenglong Ma}, \bibinfo{person}{Radha
  Chitta}, \bibinfo{person}{Jing Zhou}, \bibinfo{person}{Quanzeng You},
  \bibinfo{person}{Tong Sun}, {and} \bibinfo{person}{Jing Gao}.}
  \bibinfo{year}{2017}\natexlab{}.
\newblock \showarticletitle{Dipole: Diagnosis prediction in healthcare via
  attention-based bidirectional recurrent neural networks}. In
  \bibinfo{booktitle}{\emph{Proceedings of the 23rd ACM SIGKDD international
  conference on knowledge discovery and data mining}}.
  \bibinfo{pages}{1903--1911}.
\newblock


\bibitem[\protect\citeauthoryear{Mikolov, Chen, Corrado, and Dean}{Mikolov
  et~al\mbox{.}}{2013a}]%
        {mikolov2013efficient}
\bibfield{author}{\bibinfo{person}{Tomas Mikolov}, \bibinfo{person}{Kai Chen},
  \bibinfo{person}{Greg Corrado}, {and} \bibinfo{person}{Jeffrey Dean}.}
  \bibinfo{year}{2013}\natexlab{a}.
\newblock \showarticletitle{Efficient estimation of word representations in
  vector space}.
\newblock \bibinfo{journal}{\emph{arXiv preprint arXiv:1301.3781}}
  (\bibinfo{year}{2013}).
\newblock


\bibitem[\protect\citeauthoryear{Mikolov, Sutskever, Chen, Corrado, and
  Dean}{Mikolov et~al\mbox{.}}{2013b}]%
        {mikolov2013distributed}
\bibfield{author}{\bibinfo{person}{Tomas Mikolov}, \bibinfo{person}{Ilya
  Sutskever}, \bibinfo{person}{Kai Chen}, \bibinfo{person}{Greg~S. Corrado},
  {and} \bibinfo{person}{Jeff Dean}.} \bibinfo{year}{2013}\natexlab{b}.
\newblock \showarticletitle{Distributed representations of words and phrases
  and their compositionality}. In \bibinfo{booktitle}{\emph{Advances in Neural
  Information Processing Systems}}. \bibinfo{pages}{3111--3119}.
\newblock


\bibitem[\protect\citeauthoryear{Mullenbach, Wiegreffe, Duke, Sun, and
  Eisenstein}{Mullenbach et~al\mbox{.}}{2018}]%
        {mullenbach2018explainable}
\bibfield{author}{\bibinfo{person}{James Mullenbach}, \bibinfo{person}{Sarah
  Wiegreffe}, \bibinfo{person}{Jon Duke}, \bibinfo{person}{Jimeng Sun}, {and}
  \bibinfo{person}{Jacob Eisenstein}.} \bibinfo{year}{2018}\natexlab{}.
\newblock \showarticletitle{Explainable prediction of medical codes from
  clinical text}.
\newblock \bibinfo{journal}{\emph{arXiv preprint arXiv:1802.05695}}
  (\bibinfo{year}{2018}).
\newblock


\bibitem[\protect\citeauthoryear{Paszke, Gross, Massa, Lerer, Bradbury, Chanan,
  Killeen, Lin, Gimelshein, Antiga, et~al\mbox{.}}{Paszke
  et~al\mbox{.}}{2019}]%
        {paszke2019pytorch}
\bibfield{author}{\bibinfo{person}{Adam Paszke}, \bibinfo{person}{Sam Gross},
  \bibinfo{person}{Francisco Massa}, \bibinfo{person}{Adam Lerer},
  \bibinfo{person}{James Bradbury}, \bibinfo{person}{Gregory Chanan},
  \bibinfo{person}{Trevor Killeen}, \bibinfo{person}{Zeming Lin},
  \bibinfo{person}{Natalia Gimelshein}, \bibinfo{person}{Luca Antiga},
  {et~al\mbox{.}}} \bibinfo{year}{2019}\natexlab{}.
\newblock \showarticletitle{PyTorch: An imperative style, high-performance deep
  learning library}. In \bibinfo{booktitle}{\emph{Advances in Neural
  Information Processing Systems}}. \bibinfo{pages}{8024--8035}.
\newblock


\bibitem[\protect\citeauthoryear{Purushotham, Meng, Che, and Liu}{Purushotham
  et~al\mbox{.}}{2017}]%
        {purushotham2017benchmark}
\bibfield{author}{\bibinfo{person}{Sanjay Purushotham},
  \bibinfo{person}{Chuizheng Meng}, \bibinfo{person}{Zhengping Che}, {and}
  \bibinfo{person}{Yan Liu}.} \bibinfo{year}{2017}\natexlab{}.
\newblock \showarticletitle{Benchmark of deep learning models on large
  healthcare {MIMIC} datasets}.
\newblock \bibinfo{journal}{\emph{arXiv preprint arXiv:1710.08531}}
  (\bibinfo{year}{2017}).
\newblock


\bibitem[\protect\citeauthoryear{Rendle, Freudenthaler, Gantner, and
  Schmidt-Thieme}{Rendle et~al\mbox{.}}{2012}]%
        {rendle2012bpr}
\bibfield{author}{\bibinfo{person}{Steffen Rendle}, \bibinfo{person}{Christoph
  Freudenthaler}, \bibinfo{person}{Zeno Gantner}, {and} \bibinfo{person}{Lars
  Schmidt-Thieme}.} \bibinfo{year}{2012}\natexlab{}.
\newblock \showarticletitle{BPR: Bayesian personalized ranking from implicit
  feedback}.
\newblock \bibinfo{journal}{\emph{arXiv preprint arXiv:1205.2618}}
  (\bibinfo{year}{2012}).
\newblock


\bibitem[\protect\citeauthoryear{Soni, Ansari, Sharma, and Soni}{Soni
  et~al\mbox{.}}{2011}]%
        {soni2011predictive}
\bibfield{author}{\bibinfo{person}{Jyoti Soni}, \bibinfo{person}{Ujma Ansari},
  \bibinfo{person}{Dipesh Sharma}, {and} \bibinfo{person}{Sunita Soni}.}
  \bibinfo{year}{2011}\natexlab{}.
\newblock \showarticletitle{Predictive data mining for medical diagnosis: An
  overview of heart disease prediction}.
\newblock \bibinfo{journal}{\emph{International Journal of Computer
  Applications}}  \bibinfo{volume}{17} (\bibinfo{year}{2011}),
  \bibinfo{pages}{43--48}.
\newblock


\bibitem[\protect\citeauthoryear{Wang, He, Wang, Feng, and Chua}{Wang
  et~al\mbox{.}}{2019}]%
        {wang2019neural}
\bibfield{author}{\bibinfo{person}{Xiang Wang}, \bibinfo{person}{Xiangnan He},
  \bibinfo{person}{Meng Wang}, \bibinfo{person}{Fuli Feng}, {and}
  \bibinfo{person}{Tat-Seng Chua}.} \bibinfo{year}{2019}\natexlab{}.
\newblock \showarticletitle{Neural graph collaborative filtering}. In
  \bibinfo{booktitle}{\emph{Proceedings of the 42nd International ACM SIGIR
  Conference on Research and Development in Information Retrieval}}.
  \bibinfo{pages}{165--174}.
\newblock


\bibitem[\protect\citeauthoryear{Wang, Yang, Wen, Chen, Huang, and Zheng}{Wang
  et~al\mbox{.}}{2021}]%
        {wang2021lifelong}
\bibfield{author}{\bibinfo{person}{Zifeng Wang}, \bibinfo{person}{Yifan Yang},
  \bibinfo{person}{Rui Wen}, \bibinfo{person}{Xi Chen},
  \bibinfo{person}{Shao-Lun Huang}, {and} \bibinfo{person}{Yefeng Zheng}.}
  \bibinfo{year}{2021}\natexlab{}.
\newblock \showarticletitle{Lifelong Learning based Disease Diagnosis on
  Clinical Notes}. In \bibinfo{booktitle}{\emph{Proceedings of the 25th
  Pacific-Asia Conference on Knowledge Discovery and Data Mining (PAKDD)}}.
\newblock


\bibitem[\protect\citeauthoryear{Wei and Eickhoff}{Wei and Eickhoff}{2018}]%
        {wei2018embedding}
\bibfield{author}{\bibinfo{person}{Xing Wei} {and} \bibinfo{person}{Carsten
  Eickhoff}.} \bibinfo{year}{2018}\natexlab{}.
\newblock \showarticletitle{Embedding electronic health records for clinical
  information retrieval}.
\newblock \bibinfo{journal}{\emph{arXiv preprint arXiv:1811.05402}}
  (\bibinfo{year}{2018}).
\newblock


\bibitem[\protect\citeauthoryear{Weng, Huang, and Han}{Weng
  et~al\mbox{.}}{2016}]%
        {weng2016disease}
\bibfield{author}{\bibinfo{person}{Cheng-Hsiung Weng}, \bibinfo{person}{Tony
  Cheng-Kui Huang}, {and} \bibinfo{person}{Ruo-Ping Han}.}
  \bibinfo{year}{2016}\natexlab{}.
\newblock \showarticletitle{Disease prediction with different types of neural
  network classifiers}.
\newblock \bibinfo{journal}{\emph{Telematics and Informatics}}
  \bibinfo{volume}{33} (\bibinfo{year}{2016}), \bibinfo{pages}{277--292}.
\newblock


\bibitem[\protect\citeauthoryear{Ying, He, Chen, Eksombatchai, Hamilton, and
  Leskovec}{Ying et~al\mbox{.}}{2018}]%
        {ying2018graph}
\bibfield{author}{\bibinfo{person}{Rex Ying}, \bibinfo{person}{Ruining He},
  \bibinfo{person}{Kaifeng Chen}, \bibinfo{person}{Pong Eksombatchai},
  \bibinfo{person}{William~L. Hamilton}, {and} \bibinfo{person}{Jure
  Leskovec}.} \bibinfo{year}{2018}\natexlab{}.
\newblock \showarticletitle{Graph convolutional neural networks for web-scale
  recommender systems}. In \bibinfo{booktitle}{\emph{Proceedings of the 24th
  ACM SIGKDD International Conference on Knowledge Discovery and Data Mining}}.
  \bibinfo{pages}{974--983}.
\newblock


\bibitem[\protect\citeauthoryear{Zhang, Kowsari, Harrison, Lobo, and
  Barnes}{Zhang et~al\mbox{.}}{2018}]%
        {zhang2018patient2vec}
\bibfield{author}{\bibinfo{person}{Jinghe Zhang}, \bibinfo{person}{Kamran
  Kowsari}, \bibinfo{person}{James~H. Harrison}, \bibinfo{person}{Jennifer~M.
  Lobo}, {and} \bibinfo{person}{Laura~E. Barnes}.}
  \bibinfo{year}{2018}\natexlab{}.
\newblock \showarticletitle{{Patient2Vec}: A personalized interpretable deep
  representation of the longitudinal electronic health record}.
\newblock \bibinfo{journal}{\emph{IEEE Access}}  \bibinfo{volume}{6}
  (\bibinfo{year}{2018}), \bibinfo{pages}{65333--65346}.
\newblock


\bibitem[\protect\citeauthoryear{Zhang, Qian, Li, Liu, Chen, Guan, Zheng, and
  Li}{Zhang et~al\mbox{.}}{2021}]%
        {lixian2021}
\bibfield{author}{\bibinfo{person}{Xianli Zhang}, \bibinfo{person}{Buyue Qian},
  \bibinfo{person}{Yang Li}, \bibinfo{person}{Yang Liu}, \bibinfo{person}{Xi
  Chen}, \bibinfo{person}{Chong Guan}, \bibinfo{person}{Yefeng Zheng}, {and}
  \bibinfo{person}{Chen Li}.} \bibinfo{year}{2021}\natexlab{}.
\newblock \showarticletitle{Learning Robust Patient Representations from
  Multi-modal Electronic Health Records: A Supervised Deep Learning Approach}.
  In \bibinfo{booktitle}{\emph{Proceedings of the 2021 SIAM International
  Conference on Data Mining (SDM)}}.
\newblock


\bibitem[\protect\citeauthoryear{Zhu, Yin, Qian, Cheng, Wei, and Wang}{Zhu
  et~al\mbox{.}}{2016}]%
        {zhu2016measuring}
\bibfield{author}{\bibinfo{person}{Zihao Zhu}, \bibinfo{person}{Changchang
  Yin}, \bibinfo{person}{Buyue Qian}, \bibinfo{person}{Yu Cheng},
  \bibinfo{person}{Jishang Wei}, {and} \bibinfo{person}{Fei Wang}.}
  \bibinfo{year}{2016}\natexlab{}.
\newblock \showarticletitle{Measuring patient similarities via a deep
  architecture with medical concept embedding}. In
  \bibinfo{booktitle}{\emph{IEEE 16th International Conference on Data
  Mining}}. \bibinfo{pages}{749--758}.
\newblock


\end{thebibliography}
\end{document}